\documentclass[manuscript,screen,nonacm,preprint]{acmart}
\PassOptionsToPackage{unicode}{hyperref}
\PassOptionsToPackage{dvipsnames,svgnames,x11names}{xcolor}

\IfFileExists{bookmark.sty}{\usepackage{bookmark}}{\usepackage{hyperref}}

\newif\ifdraft
\drafttrue

\usepackage{longtable,booktabs,array}
\usepackage{calc} % for calculating minipage widths
\usepackage{etoolbox}
\usepackage{tcolorbox}
\makeatletter
\patchcmd\longtable{\par}{\if@noskipsec\mbox{}\fi\par}{}{}
\makeatother
\IfFileExists{footnotehyper.sty}{\usepackage{footnotehyper}}{\usepackage{footnote}}
\makesavenoteenv{longtable}
\usepackage{graphicx}
\makeatletter
\def\maxwidth{\ifdim\Gin@nat@width>\linewidth\linewidth\else\Gin@nat@width\fi}
\def\maxheight{\ifdim\Gin@nat@height>\textheight\textheight\else\Gin@nat@height\fi}
\makeatother
\setkeys{Gin}{width=\maxwidth,height=\maxheight,keepaspectratio}

\usepackage{booktabs}
\usepackage{longtable}
\usepackage{array}
\usepackage{multirow}
\usepackage{wrapfig}
\usepackage{float}
\usepackage{colortbl}
\usepackage{pdflscape}
\usepackage{tabu}
\usepackage{makecell}
\usepackage{xcolor}
\definecolor{mypink}{RGB}{219, 48, 122}
\makeatletter
\@ifpackageloaded{caption}{}{\usepackage{caption}}
\AtBeginDocument{%
\ifdefined\contentsname
  \renewcommand*\contentsname{Table of contents}
\else
  \newcommand\contentsname{Table of contents}
\fi
\ifdefined\listfigurename
  \renewcommand*\listfigurename{List of Figures}
\else
  \newcommand\listfigurename{List of Figures}
\fi
\ifdefined\listtablename
  \renewcommand*\listtablename{List of Tables}
\else
  \newcommand\listtablename{List of Tables}
\fi
\ifdefined\figurename
  \renewcommand*\figurename{Figure}
\else
  \newcommand\figurename{Figure}
\fi
\ifdefined\tablename
  \renewcommand*\tablename{Table}
\else
  \newcommand\tablename{Table}
\fi
}
\@ifpackageloaded{float}{}{\usepackage{float}}
\floatstyle{ruled}
\@ifundefined{c@chapter}{\newfloat{codelisting}{h}{lop}}{\newfloat{codelisting}{h}{lop}[chapter]}
\floatname{codelisting}{Listing}

\makeatother
\makeatletter
\@ifpackageloaded{caption}{}{\usepackage{caption}}
\@ifpackageloaded{subcaption}{}{\usepackage{subcaption}}
\makeatother
\makeatletter
\@ifundefined{shadecolor}{\definecolor{shadecolor}{rgb}{.97, .97, .97}}
\makeatother
\AtBeginDocument{%
  }

\copyrightyear{2024}
\acmYear{2024}
\setcopyright{rightsretained}
\acmConference[FAccT '24]{The 2024 ACM Conference on Fairness, Accountability, and Transparency}{June 3--6, 2024}{Rio de Janeiro, Brazil}
\acmBooktitle{The 2024 ACM Conference on Fairness, Accountability, and Transparency (FAccT '24), June 3--6, 2024, Rio de Janeiro, Brazil}\acmDOI{10.1145/3630106.3658974}
\acmISBN{979-8-4007-0450-5/24/06}

\author{Jan Simson}
\affiliation{%
  \institution{LMU Munich}
  \city{Munich}
  \country{Germany}}
\affiliation{%
  \institution{Munich Center for Machine Learning (MCML)}
  \city{Munich}
  \country{Germany}}
\email{jan.simson@lmu.de}

\author{Florian Pfisterer}
\affiliation{%
  \institution{LMU Munich}
  \city{Munich}
  \country{Germany}}

\author{Christoph Kern}
\affiliation{%
  \institution{LMU Munich}
  \city{Munich}
  \country{Germany}}
\affiliation{%
  \institution{Munich Center for Machine Learning (MCML)}
  \city{Munich}
  \country{Germany}}
\affiliation{%
  \institution{University of Maryland}
  \city{College Park}
  \country{USA}}
\email{christoph.kern@lmu.de}

\begin{document}

\title[One Model Many Scores]{One Model Many Scores: Using Multiverse Analysis to Prevent Fairness Hacking and Evaluate the Influence of Model Design Decisions}


\begin{abstract}

A vast number of systems across the world use algorithmic decision making (ADM) to (partially) automate decisions that have previously been made by humans. The downstream effects of ADM systems critically depend on the decisions made during a systems' design, implementation, and evaluation, as biases in data can be mitigated or reinforced along the modeling pipeline. Many of these decisions are made implicitly, without knowing exactly how they will influence the final system. To study this issue, we draw on insights from the field of psychology and introduce the method of multiverse analysis for algorithmic fairness. In our proposed method, we turn implicit decisions during design and evaluation into explicit ones and demonstrate their fairness implications. By combining decisions, we create a grid of all possible ``universes'' of decision combinations. For each of these universes, we compute metrics of fairness and performance. Using the resulting dataset, one can investigate the variability and robustness of fairness scores and see how and which decisions impact fairness. We demonstrate how multiverse analyses can be used to better understand fairness implications of design and evaluation decisions using an exemplary case study of predicting public health care coverage for vulnerable populations. Our results highlight how decisions regarding the evaluation of a system can lead to vastly different fairness metrics for the same model. This is problematic, as a nefarious actor could optimise or ``hack'' a fairness metric to portray a discriminating model as fair merely by changing how it is evaluated. We illustrate how a multiverse analysis can help to address this issue.

\end{abstract}




\maketitle

\begin{tcolorbox}[colframe=red, colback=white, arc=2mm, boxrule=1mm, width=\textwidth]

\textbf{When referencing this work, please cite it as follows:}

Jan Simson, Florian Pfisterer, and Christoph Kern. 2024. One Model Many Scores: Using Multiverse Analysis to Prevent Fairness Hacking
and Evaluate the Influence of Model Design Decisions. In \textit{The 2024 ACM Conference on Fairness, Accountability, and Transparency (FAccT ’24), June 3–6, 2024, Rio de Janeiro, Brazil}. ACM, New York, NY, USA, 25 pages. \href{https://doi.org/10.1145/3630106.3658974}{https://doi.org/10.1145/3630106.3658974}

\end{tcolorbox}

\setlength{\parskip}{-0.1pt}

\hypertarget{introduction}{%
\section{Introduction}\label{introduction}}

Across the world, more and more decisions are being made with the
support of machine learning (ML) and algorithms; so called algorithmic
decision making (ADM). Examples of such systems can be found in finance
for loan approvals \citep{mukerjee2002}, the labor market for hiring
decisions or filtering resumes \citep{faliagka2012}, and the criminal
justice system to assess risks of recidivism \citep{angwin2016}. While
these systems are promising when designed well, raising hopes of
more accurate and objective decisions, their impact can be quite the
opposite when designed incorrectly. There are many examples of ADM systems
discriminating against people \citep{mehrabi2021}. One prominent example
was the \emph{robodebt} system, where the Australian government used an
algorithm to detect potential social security overpayments. Due to
serious flaws in the design of the system, it often overestimated debts
and put the burden on the accused to prove the contrary
\citep{henriques-gomes2023}. Other examples include the Dutch childcare
benefits system using an ADM system that was much more likely to accuse
immigrants of having committed fraud \citep{amnestyinternational2021}.

These fairness problems often occur because algorithms replicate biases
in the underlying training data. However, biases can also be amplified
throughout the machine learning pipeline depending on how exactly data
is processed and turned into outputs \citep{kern, rodolfa2020}.
Unfortunately, no silver bullet exists to prevent biases
in the machine learning pipeline \citep{agrawal} and legislation usually
provides little guidance. Understanding how modeling decisions interact
with fairness is therefore a prerequisite for effectively mitigating
unintended outcomes in practice. A systematic mapping of design
decisions to fairness outcomes can critically guide the model selection
process, as multiple models may achieve similar accuracy, but can
considerably differ in their fairness properties \citep{black2022}. Alarmingly, we demonstrate how the evaluation of the same model can
be modified to achieve large variability in a fairness metric,
potentially allowing the \emph{hacking} of fairness metrics. Related issues
regarding the hacking or washing of fairness metrics have recently been raised
in fair ML research \citep{meding2024fairness, aivodji2019fairwashing}.
As a result, preventing algorithms from introducing, reinforcing or hiding
biases requires careful study and evaluation of the -- often implicit --
decisions made while designing and evaluating a machine learning system.
To address this objective in a systematic and efficient way, we
introduce the method of multiverse analysis for algorithmic fairness.
Multiverse analyses were introduced to psychology with the intent to
improve reproducibility and create more robust research
\citep{steegen2016}. We adapt this methodology across domains to work in
the context of machine learning with a focus on evaluating metrics of
algorithmic fairness. We present two variations of this method
demonstrating its usefulness: (1) as a guidance during the design of the
model and preprocessing pipeline and (2) as an estimator of robustness
of a fairness metric and to protect against fairness hacking.

\begin{figure}

{\centering \includegraphics[width=0.7\textwidth]{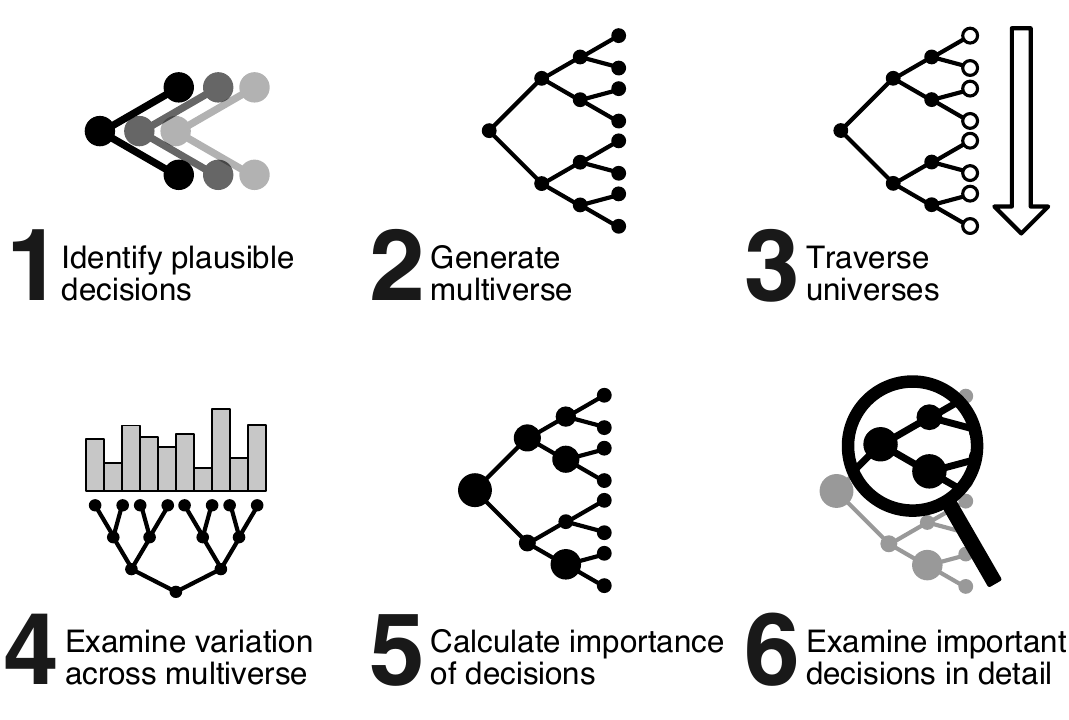}

}

\caption{\label{fig-method-explanation}\textbf{Steps to conduct a
multiverse analysis for algorithmic fairness.} Steps 1 - 4 apply to
multiverse analyses in general, whereas steps 5 - 6 are unique to
larger multiverse analyses for algorithmic fairness.}

\end{figure}

In the following, we present a generalizable approach of using
multiverse analysis to estimate the effect of decisions during the
design and evaluation of a machine learning or ADM system on fairness
outcomes. Using a case study of predicting public health coverage in US
census data we demonstrate how design decisions can be better understood
and fairness hacking can be addressed. We provide modular source code to
allow streamlined adaptation of the proposed method in other use cases
and contexts.

\hypertarget{multiverse-analysis}{%
\subsection{Multiverse Analysis}\label{multiverse-analysis}}

Multiverse analyses were first introduced in psychology by
\citet{steegen2016} in response to the reproducibility crisis affecting
the field \citep{opensciencecollaboration2015}. The goal of this
analysis type is to investigate the invariance of results to
researchers' analysis decisions. Specifically, when analyzing a dataset,
researchers make many implicit and explicit choices \citep{simmons2011},
often without the option of confirming whether a choice is correct or
incorrect. This leads to many plausible scenarios when analyzing data,
as one traverses a \emph{garden of forking paths} \citep{gelman2014},
where each fork corresponds to a decision. The multitude of these
scenarios becomes especially evident when multiple researchers analyze
the same data, coming to staggeringly different results
\citep{breznau2022}.

Multiverse analysis focuses on the preprocessing steps applied to a
dataset: Steps such as selecting the observations and predictor
variables to include in a dataset or scaling and binning their values.
Based on the different decisions made and paths taken when
preprocessing a dataset, analysts will end up with one of many possible
datasets for the actual analysis. In a multiverse analysis, the goal is
to make this variation explicit by using the complete grid of decisions
and their options to generate all plausible datasets. Using all
potential datasets, a multiverse analysis re-runs the analysis on each
of them to receive the distribution of results instead of a single
result point (Figure~\ref{fig-method-explanation}, Steps 1 - 3). We extend
this methodology to also examine the influence of variation in evaluation
and adapt it for the machine learning context with a special
focus on using it to generate insights on metrics of algorithmic
fairness.

In addition to multiverse analysis, a related type of analysis, called specification curve analysis \citep{simonsohn2020specification} emerged in the social sciences literature. Its goal is to assess the strength of an effect of interest under the different modelling decisions contained in the complete grid of possible decision combinations. Results are aggregated in a specification curve, a graph displaying the distribution of the effect size or coefficient of interest, yielding a single curve that allows assessing the robustness of a measured association across modelling decisions. In contrast, our approach is not only interested in the robustness, but we aim to also identify decisions that impact the resulting fairness metrics for further investigation.

\hypertarget{multiverse-analysis-for-algorithmic-fairness}{%
\subsection{Multiverse Analysis for Algorithmic
Fairness}\label{multiverse-analysis-for-algorithmic-fairness}}

In our proposed adaptation of multiverse analysis for algorithmic
fairness, one starts by compiling a list of all potentially relevant
decisions that are being made during the design and evaluation of a particular system.
We differentiate between different kinds of decisions in this context:
(1) decisions which are already made explicitly with a consideration of
their different options e.g.~choice of model and its hyperparameters,
and (2) decisions which are made explicitly, but without any
consideration for alternatives e.g.~log-transforming an income column
because it is common practice. In a multiverse analysis, the goal is to
turn both types of decisions into completely explicitly made decisions
and evaluate their impacts. There are also decisions which may initially
not even be considered as such e.g.~modifying classification cutoffs
post-hoc due to external constraints. Conducting a multiverse analysis
invites reflection on the modeling pipeline such that implicit decisions
may surface and are turned into explicit ones. One of the key
differences in the present analysis compared to a classic multiverse
analysis is that we will evaluate machine learning systems, whereas
classical multiverse analyses will typically evaluate the outcomes of
null-hypothesis-significance-tests (NHST) across analysis choices. While
many of the decision points apply to any machine learning system (e.g.,
choice of algorithm, how to preprocess certain variables,
cross-validation splits), many of them are also domain-specific (e.g.,
coding of certain variables, how to set classification thresholds, how
fairness is operationalized). We focus on decisions made during the
preprocessing of data, in line with the original approach of multiverse
analyses \citep{steegen2016}. We extend this approach to incorporate
decisions relevant to algorithmic fairness, particularly with regard to
protected attributes and the translation of predictions into real-world
actions or interventions. Similarly to a classical multiverse analysis,
we use the resulting \emph{garden of forking paths} to generate a grid
of all possible universes of decision combinations, the multiverse. For each of these universes, we compute the resulting fairness and performance metrics of the machine learning system and collect them as a data point. Based on the
resulting dataset of decision universes and corresponding fairness
scores, we evaluate how individual decisions influence the fairness
metric and explore the most important decisions in more detail
(Figure~\ref{fig-method-explanation}).

Another novelty in our approach is our introduction of two distinct
perspectives on multiverse analyses: One with a focus on preprocessing,
fostering the understanding of how decisions affect models in a fairness
context and a second, focusing on robust fairness evaluation of ML
systems and protecting against cherry picking of evaluation criteria.

\hypertarget{related-research}{%
\subsection{Related Research}\label{related-research}}

Existing work has described the effects of specific preprocessing or modeling decisions in isolation, such as the influence of different imputation methods \citep{caton2022}, of the model architecture, and of hyperparameters \citep{dooley2024rethinking} on fairness in different contexts.
Multiverse analyses have also been used to model the performance distribution in hyperparameter-space \citep{bell}, but not yet to analyze algorithmic fairness. Research into model multiplicity has discovered multiple sources of arbitrariness that can influence model predictions and fairness: Random samples of a dataset can lead to different predictions on the individual level \citep{cooper2024arbitrariness,friedler2019comparative}, the selection of different target variables can strongly affect model fairness \citep{watson-daniels2023multitarget} and even the original sampling during the creation of a dataset can be considered arbitrary \citep{meyer2023dataset}.

In terms of manipulating fairness, prior work has demonstrated the possibility of generating surrogate models that show little dependence on protected features for unfair models, a process termed ``fairwashing'' \citep{aivodji2019fairwashing}. Under an assumption of ``fairness through unawareness'', these surrogate models could then be presented as fair models. This assumption is unrealistic in practice, however, as there are commonly proxy variables available for protected attributes \citep{barocas2023classification}. Recent parallel work has demonstrated a process of using completely different fairness metrics to then report only the one with the most optimal score in a process also termed ``fairness hacking'' \citep{meding2024fairness}. In this work, we demonstrate how there is no need to vary the chosen fairness metric itself, if one is willing to shift evaluation criteria in order to manipulate its scores. We believe both of these approaches are troublesome and fall under the term ``fairness hacking''. They closely mirror practices of varying evaluation criteria to achieve significant p-values, a practice commonly referred to as ``p-hacking'', which gave rise to the introduction of multiverse analysis in psychology in the first place \citep{simonsohn2014pcurve}.

The field of hyperparameter-optimization (HPO)
\citep{feurer2019, bischl2023} tries to optimize the process of tuning
machine learning model hyperparameters. This field typically focuses on
optimizing algorithm performance by employing efficient search
strategies that allow optimizing performance
without requiring the exploration of the complete hyperparameter space.
However, adaptive search patterns such as, e.g. Bayesian Optimization \citep{snoek2012practical},  usually focus on efficiently finding the optimal
configuration and yield non-i.i.d. optimization traces. This makes them unsuitable for assessing the influence and robustness of any particular decision as post-hoc analysis relies on representative, i.i.d. data. While algorithmic fairness is also explored in the context of HPO \citep{pfisterer-arxiv19a, perrone2021}, the focus is only on finding models with favourable performance-fairness trade-offs instead of understanding the effects of individual decisions or assessing overall robustness. Here, we draw on insights and methodology from the field of HPO, in particular the functional analysis of variance (FANOVA) \citep{hooker2007, hutter2014} to allow a more interpretable and efficient analysis of the results from the multiverse analysis. Our focus, however, is on uncovering and systematically exploring variation induced by the different decisions instead of finding the setting that optimizes fairness metrics.

\hypertarget{case-study}{%
\subsection{Case Study}\label{case-study}}

We illustrate how multiverse analysis can enrich the machine learning
fairness toolkit using a case study of predicting public health
insurance coverage. Accurate and fair prediction of public health
insurance coverage in the United States is an important issue as access
to healthcare is quite expensive in the US, with the country spending
almost \(16\%\) of its gross domestic product per capita on healthcare in 2020
\citep{ortiz-ospina2017}. Whether or not someone is covered by health
insurance can have large effects on their health and financial situation:
According to \citet{sommers2017}, people with insurance have better self-reported health, have more preventative doctor's appointments, improved depression outcomes, and fewer personal bankruptcies.

We implement our case study using the ACSPublicCoverage dataset \citep{ding2021}, with data from the American Community Survey (ACS) Public Use Microdata Sample (PUMS) \citep{us2021understanding}. We use this particular dataset as it is rich enough
for us to implement a wide range of design decisions and because many
other well-established datasets used in the fairness literature suffer
from non-trivial quality issues \citep{ding2021, fabris2022, bao}: UCI
Adult \citep{kohavi1996}, the most popular dataset in the fairness
literature \citep{fabris2022}, uses an arbitrary threshold of \$50,000
to create a binary task of income prediction. This threshold has been
shown to greatly influence the accuracy of predictions in certain
groups, biasing measures of algorithmic fairness and threatening
external validity \citep{ding2021}. The ACSPublicCoverage dataset is one
of the datasets which have been specifically developed in response to
the issues in UCI Adult.

Here, we operationalize having public insurance coverage as being covered by either Medicare, Medicaid, Medical Assistance (or any kind of government-assistance plan for those with low incomes or a disability) or Veterans Affairs Health Care, following the official Guidance for Health Insurance Data Users from the US Census Bureau \citep{uscensusbureau}. In line with the original task setup by \citet{ding2021}, only individuals with an age below 65 years and a yearly income of less than \$30,000 are examined. Low-income households are also more likely to rely on public health insurance \citep{keisler-starkey2022}.

As there are no clear guidelines on how to set up an ADM system within
this context (as would be the case in heavily regulated contexts such as
credit scoring) one faces a multitude of decisions when designing a
solution for this task, each of which can govern how bias is fed into
the final system. A multiverse analysis for algorithmic fairness
requires developers to make these design decisions explicit and shows
their fairness implications in the present context.

\hypertarget{methodology}{%
\section{Methodology}\label{methodology}}

\hypertarget{fairness-metric}{%
\subsection{Fairness Metric}\label{fairness-metric}}

While our proposed analysis works with multiple different fairness
metrics, it requires one to choose a primary metric for analysis. For
the present case study we used \emph{equalized odds difference}
\citep{agarwal, hardt} as the primary fairness metric, as it quantifies
the degree to which a system's predictions are equally good across
different groups defined by a protected attribute. Equalized odds
require both the \emph{true positive rate} (TPR) and the \emph{false
positive rate} (FPR) of a system's predictions to be equal across all
groups of the protected attribute. Values of the \emph{equalized odds
difference} can range from \(0\) to \(1\). A value of \(0\) corresponds
to a perfectly fair model according to the metric, whereas a value of
\(1\) corresponds to a completely unfair model. We use the
implementation from the fairlearn package \citep{weerts2023} to calculate
the metric, where the differences in both the \emph{true positive rate}
and the \emph{false positive rate} are calculated and the larger of the
two is used as the metric. We consider \emph{race} as the protected
attribute in our case study given the persisting racial disparities in
various domains, including health outcomes, in the US
\citep{obermeyer2019} and matching the original task \citep{ding2021}.

\hypertarget{decision-space}{%
\subsection{Decision Space}\label{decision-space}}

When conducting a multiverse analysis, the first step is the
identification of relevant and plausible decisions to be made. Based on
the literature on data science and machine learning workflows
\citep{kuhn2020, lequy2022} we identified five distinct categories to
structure and guide the identification of decisions: Data Selection,
Preprocessing, Modeling, Post-Hoc and Evaluation decisions (Table
\ref{tbl-decisions}). As there is a potentially infinite list of possible decisions to consider, the present list is not intended to be exhaustive, but rather to highlight the most common and important categories of decisions one may typically encounter when designing a machine learning or ADM system. We also deliberately set the focus on decisions where alternative options are typically not considered or ones that are not identified as decisions at all. When adapting the methodology to a new system, this list can serve as an inspiration, however, one must also consider the domain-specific decisions unique to each applied problem.

\hypertarget{tbl-decisions}{}
\begin{table}
\caption{\label{tbl-decisions}Overview of the typical decision categories, the actual decisions
examined in the case study and their respective options used to
construct the multiverse. }\tabularnewline

\centering
\begin{tabular}{ll>{\raggedright\arraybackslash}p{9cm}}
\toprule
\multicolumn{1}{c}{\em{ }} & \multicolumn{2}{c}{\em{Decisions and Options Examined in Case Study}} \\
\cmidrule(l{3pt}r{3pt}){2-3}
Category & Decision & Options\\
\midrule
\addlinespace[0.3em]
\multicolumn{3}{l}{\textbf{Decisions examined in Study 1}}\\
\hspace{1em}\textit{Data Selection} & Exclude Features & (1)~none; (2)~race; (3)~sex; (4)~race-sex\\
\hspace{1em} & Exclude Subgroups & (1)~keep-all; (2)~drop-smallest-1; (3)~drop-smallest-2; (4)~keep-largest-2; (5)~drop-other\\
\hspace{1em}\textit{Preprocessing} & Scale & (1)~do-not-scale; (2)~scale\\
\hspace{1em} & Preprocess Age & (1)~none; (2)~bins-10; (3)~quantiles-3; (4)~quantiles-4\\
\hspace{1em} & Preprocess Income & (1)~none; (2)~bins-10000; (3)~quantiles-3; (4)~quantiles-4\\
\hspace{1em} & Encode Categorical & (1)~one-hot; (2)~ordinal\\
\hspace{1em}\textit{Modeling} & Model & (1)~logreg; (2)~rf; (3)~gbm; (4)~elasticnet\\
\hspace{1em} & Stratify Split & (1)~none; (2)~target; (3)~protected-attribute; (4)~both\\
\hspace{1em}\textit{Post-Hoc} & Cutoff & (1)~raw-0.5; (2)~quantile-0.1; (3)~quantile-0.25\\
\addlinespace[0.3em]
\multicolumn{3}{l}{\textbf{Decisions examined in Study 2}}\\
\hspace{1em}\textit{Evaluation} & Eval Fairness Grouping & (1)~majority-minority; (2)~separate\\
\hspace{1em} & Eval Exclude Subgroups & (1)~exclude-in-eval; (2)~keep-in-eval\\
\hspace{1em} & Eval On Subset & (1)~full; (2)~locality-largest-only; (3)~locality-most-privileged; (4)~locality-city-la; (5)~locality-city-sf; (6)~exclude-military; (7)~exclude-non-citizens\\
\bottomrule
\end{tabular}
\end{table}

We chose to examine evaluation decisions separately from preprocessing decisions to demonstrate the two main uses of a multiverse analysis for algorithmic fairness: Understanding fairness implications of design decisions during model development and studying robustness of fairness scores in model evaluations. We therefore split the list of decisions as well as the following analyses into \textit{Study 1} examining the impact of design decisions on models and \textit{Study 2} examining the variation that can arise from differences in evaluation decisions. An overview of all decisions and their respective options can
be seen in Table \ref{tbl-decisions}, and a detailed description of each
is provided below.

\hypertarget{study-1-model-design-decisions}{%
\subsubsection{Study 1: Model Design
Decisions}\label{study-1-model-design-decisions}}

We consider 9 distinct and orthogonal design
decisions. Each of these decisions has two to five unique choice
options, leading to a total of \(N=61440\) combinations of decisions or
universes. We consider decisions roughly in the order they would
be made during a typical analysis.

\hypertarget{excluding-variables-as-predictors-exclude-features}{%
\textbf{Excluding Variables as Predictors (Exclude
Features).}\label{excluding-variables-as-predictors-exclude-features}} Selecting features to train a model on presents a critical design
decision. In the ADM context, it can be required to exclude certain
protected features (such as sex/gender, race, ethnicity) as predictors
due to legal constraints when designing a machine learning system.
However, as prominently shown in various studies this does not
necessarily lead to increased fairness, as the protected attribute is
often correlated with other (``legitimate'') features \citep{weerts}. We implement the following options for this decision in our case study:
(1) use all features as predictors (incl.~protected ones), (2) exclude
race, the protected attribute in the case study, (3) exclude sex, a
sensitive attribute and (4) exclude both race and sex from modelling.

\hypertarget{sec-exclude-subgroups}{%
\textbf{Excluding Subgroups of the Protected Attribute (Exclude
Subgroups).}\label{sec-exclude-subgroups}} When working with variables with an uneven distribution or very rare
categories one may focus only on the most common groups, dropping data
for smaller ones. This can be done to preserve the privacy of small
groups, due to unreliability in the data or out of convenience to allow for an easier model interpretation
downstream. However, the exclusion of subgroups of the population can
potentially be harmful, with discriminatory differences in downstream
model predictions. While we decided to include this practice as a
decision in our analysis to (1) raise awareness of the issue and (2)
represent the effects of the practice in our analysis, this should not
be taken as an endorsement of this practice. We try to capture the implications of this practice via the attribute
race. We therefore chose to include a decision of dropping certain groups from the training data based on their prevalence. Groups were \emph{not} dropped from the test data used for evaluation as part of this decision. We include six options for this decision, with the fraction of discarded data in brackets\footnote{Fractions of discarded training data are only reported for a non-stratified train-test split, as there are only \emph{very slight} differences in the fraction of discarded data based on stratification strategy.}: (1) to keep all groups (\(0.00\%\)), (2)
to drop the smallest group (\(0.01\%\)), (3) to drop the two smallest
groups (\(0.33\%\)), (4) to keep the two largest groups (\(27.45\%\))
and (5) to drop the category ``Some Other Race alone'' specifically
(\(15.81\%\)).

\hypertarget{scaling-of-continuous-variables-scale}{%
\textbf{Scaling of Continuous Variables
(Scale).}\label{scaling-of-continuous-variables-scale}} It is common to scale continuous variables during preprocessing,
centering them on a mean of \(\mu = 0\) and standard deviation of
\(\sigma = 1\) (also referred to as z-scaling). Scaling may be
particularly advisable if kernel-based learners are used as it typically
leads to improved performance for such models. We include two options for this decision: (1) to keep continuous
variables as they are and (2) to scale continuous variables.

\hypertarget{binning-of-continuous-variables-preprocess-age-preprocess-income}{%
\textbf{Binning of Continuous Variables (Preprocess Age, Preprocess
Income).}\label{binning-of-continuous-variables-preprocess-age-preprocess-income}} Another common practice is binning continuous variables, i.e., turning
continuous variables into ordinal variables with discrete categories.
The reasons to do this are plentiful: To deal with outliers, to address
privacy concerns, or for a more tangible interpretation to name a few. We provide two distinct and orthogonal decisions here on whether or how
to bin the variables \(age\) and \(income\). We include four options for
the variable \(age\): (1) perform no binning, (2) bin into bins of size
\(10\), (3) bin into three evenly sized quantiles, (4) bin into four
evenly sized quantiles. Likewise, we include four options for the
variable \(income\): (1) perform no binning, (2) bin into bins of size
\(10,000\), (3) bin into three evenly sized quantiles, (4) bin into four
evenly sized quantiles.

\hypertarget{encoding-of-categorical-variables-encode-categorical}{%
\textbf{Encoding of Categorical Variables (Encode
Categorical).}\label{encoding-of-categorical-variables-encode-categorical}} Another common preprocessing step includes transforming categorical
variables into a numerical format. When doing this, one typically has two
options: (1) One-hot (or dummy) coding each variable with \(K\)
categories into \(K\) (or \(K - 1\)) new binary variables or (2)
ordinally encoding each variable by assigning an integer value from
\(1\) to \(K\) for each category. Ordinal encoding is only applicable,
however, for variables with a natural ordering. For all ordinal variables (including continuous variables that have been
binned), we include both options. Any variables without a natural
ordering are always one-hot coded.

\hypertarget{model-type-model}{%
\textbf{Model Type (Model).}\label{model-type-model}} A major choice when designing any statistical or machine learning system is which model type one decides to use. While there is a large number of potential models to explore here, we focused on the most commonly used ones in the context of ADM in the literature. We note that hyperparameter selection has shown to have an impact on fairness, but choose to focus on other choices, as HPO has already been studied elsewhere \citep{perrone2021}. We therefore support the following model types as options for this
decision: (1) logistic regression \citep{cox1958}, (2) random forest
\citep{ho1995}, (3) gradient boosting machine \citep{friedman2001}, and
(4) elastic net \citep{zou2005} trained with their default hyperparameters.

\hypertarget{stratification-of-train-test-split-stratify-split}{%
\textbf{Stratification of Train-Test Split (Stratify
Split).}\label{stratification-of-train-test-split-stratify-split}} Training and test sets are often created by simple random splitting of
the full dataset. It can be beneficial, however, to perform this split
conditional on certain groupings to ensure equal representation of all
labels within both the train and test sets. We include four options for this decision: (1) to not stratify at all,
using a completely random split instead, (2) to stratify using the
target variable (\emph{public coverage}), (3) to stratify using the
protected attribute (\emph{race}) and (4) to stratify using a
combination of both variables.

\hypertarget{cutoff-for-final-classification-cutoff}{%
\textbf{Cutoff for Final Classification
(Cutoff).}\label{cutoff-for-final-classification-cutoff}} At the end of the ML pipeline, the prediction models' (risk) scores can
be used to classify new observations based on a pre-specified
classification threshold. By default a threshold of \(0.5\) would be
used with every score equal or above classified as \(1\) (\emph{having
coverage}) and everything below as \(0\) (\emph{not having coverage}).
Actual interventions, however, are often based on the ranked list of
scores such that (costly) interventions are targeted at the top \(X\)
percent with the highest risk. With real-world scenarios often coming
with resource-bound restrictions, one may for example only be able to
provide an intervention for, say, \(10\%\) or \(25\%\) of the most
in-need in the population. These real-world restrictions are typically
not taken into account in fairness evaluations, despite having
potentially devastating implications. We therefore also consider different cutoff values for the final
predictions of the system. We support the following options for this
decision: (1) use the default raw cutoff value of \(0.5\), (2) only
treat the lowest \(0.1\) quantile as \emph{not having coverage}, (2)
only treat the lowest \(0.25\) quantile as \emph{not having coverage}.

\hypertarget{study-2-evaluation}{%
\subsubsection{Study 2: Evaluation}\label{study-2-evaluation}}

We consider 3 distinct and orthogonal decisions, all focusing on evaluation only. Each decision has between 2 and 7 options each. Together these produce a total of $N = 28$ unique evaluation strategies for any given model, without modifying the model or its predictions.

\hypertarget{grouping-of-protected-attribute-fairness-grouping}{%
\textbf{Grouping of Protected Attribute (Fairness
Grouping).}\label{grouping-of-protected-attribute-fairness-grouping}} When working with a fairness metric, it is necessary to specify for
which groups of the protected attribute it is calculated. The present
case study uses \emph{race} as the protected attribute. For protected
attributes with more than two categories, however, multiple comparisons
can be computed. Depending on the application context one may, e.g.,
simplify these groups into the largest group (\emph{majority}) and all
other groups (\emph{minority})\footnote{\textbf{Majority group}: `White
  alone'; \textbf{Minority group(s)}: `Asian alone', `Two or More Races',
  `Some Other Race alone', `Black or African American alone', `American
  Indian alone', `Native Hawaiian and Other Pacific Islander alone',
  `American Indian and Alaska Native tribes specified; or American
  Indian or Alaska Native, not specified and no other races' and `Alaska
  Native alone'.}. An important note regarding this decision is that it changes how the
fairness metric is calculated: with two groups, the difference between
those two groups is calculated, however, with more than two groups all
possible differences between group-pairs are calculated and the largest
difference between them is used (the default behaviour in \citet{weerts2023}). Naturally, this has a strong influence on the fairness metric.
 We include two options for this decision: (1) The fairness metric is
computed between the \emph{majority} group and \emph{minority} group and
(2) the fairness metric is computed as the maximum of the metric as
computed between all groups of the protected attribute
(\emph{race}).

\hypertarget{exclusion-of-subgroups-during-evaluation-eval-exclude-subgroups}{%
\textbf{Exclusion of Subgroups during Evaluation (Eval Exclude
Subgroups).}\label{exclusion-of-subgroups-during-evaluation-eval-exclude-subgroups}} Similarly to how subgroups of the protected attribute may be excluded from the training data, they
may also be excluded from the test data used for evaluation, with
potentially even greater adverse impact. We examine the exclusion of the same subgroups as in the decision
\emph{Exclude Subgroups} in Study 1
(Section~\ref{sec-exclude-subgroups}) and vary whether or not subgroups
are also excluded from the test dataset. The same warnings raised for
that decision are even more relevant for this decision and we
\emph{strongly} discourage the exclusion of subgroups in any system.

\hypertarget{evaluation-using-a-subset-of-the-data-eval-on-subset}{%
\textbf{Evaluation using a Subset of the Data (Eval on Subset).}\label{evaluation-using-a-subset-of-the-data-eval-on-subset}} When assessing the fairness of a system, the evaluation may happen on only a subset of the eventual target population, for example because some populations may be easier to reach or because the model deployment context changes over time. While this practice is obviously not desirable, it may be necessary in certain situations due to real-world limitations in resources. An example of this is the popular COMPAS dataset \citep{angwin2016} which was constructed using only data from a single county (Broward County, Florida), as a larger-scale construction of such a dataset would not have been feasible. We examine the following options for this decision, to represent possible population subsets one may use for evaluation: (1) examining only the largest geographical region (in terms of sample size), (2) examining the geographical region with the largest fraction of the privileged group; examining only data from the counties of (3) Los Angeles or (4) San Francisco, (5) examining a subset of only non-military people (as former military status may affect healthcare status), (6) examining only U.S. citizens and (7) not examining any subset, but rather using the full test data for evaluation.

\hypertarget{technology}{%
\subsection{Software}\label{technology}}

Analyses were conducted using Python Version 3.8 \citep{vanrossum2009}
and pipenv \citep{pipenvmaintainerteam2017} for reproducibility. The
Python package \mbox{scikit-learn} \citep{pedregosa2011} was used for
preprocessing and fitting of models, pandas \citep{team2020} for
loading and modification of data, folktables \citep{ding2021} for
retrieval of data, fairlearn \citep{weerts2023} for computation of
fairness metrics, fANOVA \citep{hutter2014} for calculation of variable
importance and papermill \citep{contributors2017} for parameterized
computation of decision universes. This reproducible document was
generated using quarto \citep{allaire2022}, R \citep{rcoreteam2022}
Version 4.2, the R packages from the tidyverse \citep{wickham2019} and
ggpubr \citep{kassambara2023} for generation of figures. The source code
of the analyses and this publication is available at
\mbox{\url{https://github.com/reliable-ai/fairml-multiverse}}. We purposefully created source code in a modular fashion to allow for easy adoption of the multiverse method in other fair ML contexts. An interactive analysis of a subset of the results is available at \url{https://reliable-ai.github.io/fairml-multiverse/}.

\hypertarget{results}{%
\section{Results}\label{results}}

\hypertarget{study-1-model-design-decisions-1}{%
\subsection{Study 1: Model Design}\label{study-1-model-design-decisions-1}}

\begin{figure}

{\centering \includegraphics[width=0.6\textwidth]{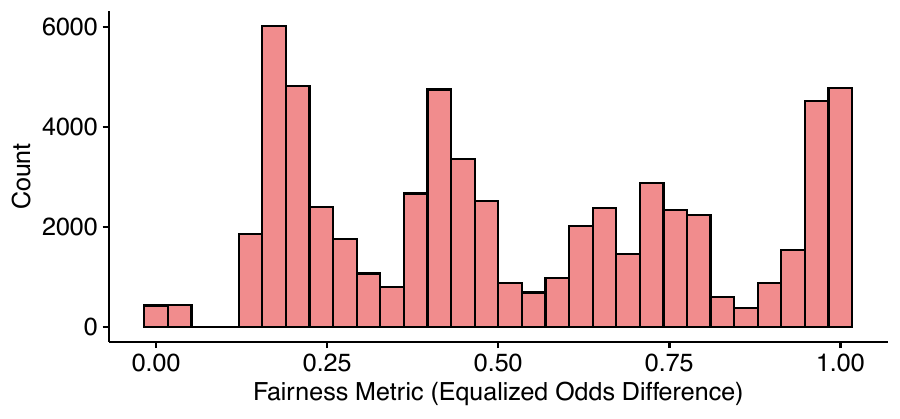}

}

\caption{\label{fig-variance}\textbf{Variation in the multiverse spans
the entirety of possible values of the fairness metric.} Distribution of
fairness metric (equalized odds difference) across universes. Lower
values on the fairness metric indicate smaller \emph{TPR} and \emph{FPR}
differences across groups.}

\end{figure}

\begin{figure}

{\centering \includegraphics[width=0.7\textwidth]{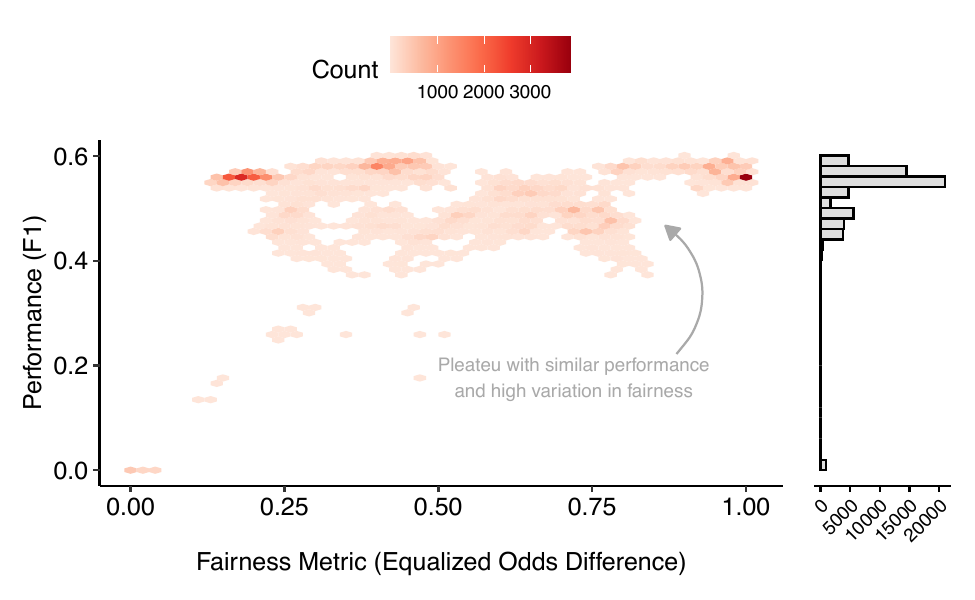}

}

\caption{\label{fig-performance-fairness}\textbf{Performance and
fairness are largely unrelated with plateaus of low variance in
performance, but high variance in fairness.} Distribution of overall
performance as \(F_{1}\) score and fairness metric (equalized odds
difference) across all multiverses. Marginal histogram shows
distribution of performance. A marginal histogram of the fairness metric
can be seen in Figure~\ref{fig-variance}, similar figures for raw and
balanced accuracy can be seen in Figure~\ref{fig-performance-fairness-acc}. An \href{https://reliable-ai.github.io/fairml-multiverse/}{interactive version} of this figure is available.}

\end{figure}

The multiverse analysis examining the influence of model design decisions produced a total of \(N = 61440\) values of the fairness metric in Study 1\footnote{In Study 1, we evaluated all models using the same strategy, namely not aggregating groups of the protected attribute, not excluding any subgroups during evaluation, and evaluating on the complete test set.\label{fn1eval}}. When examining the distribution of the fairness metric across the multiverse of decisions, the large variation of the fairness metric becomes apparent, with values spanning the entire possible range of the metric from \(0\) to \(1\)
(Figure~\ref{fig-variance}). Overall performance of the resulting models
was moderate with \(F_{1}\) scores between \(0\) and \(0.598\) and raw
accuracies between \(0.419\) and \(0.722\). Performance and the fairness
metric were only weakly correlated with a Pearson correlation of
\(r = 0.149\) for \(F_{1}\) scores and \(r = 0.192\) for raw accuracy.
For the \(F_{1}\) score, the majority of universes fell into a similar
range of performance, but exhibited large variation on the fairness
metric (Figure~\ref{fig-performance-fairness}), highlighting the
opportunity to optimize algorithmic fairness without sacrificing
performance in line with \citet{islam2021}. Raw accuracy exhibited
similar opportunities, varying largely based on the decision
\emph{Cutoff}, with three large clusters of similar performance
(Figure~\ref{fig-performance-fairness-acc} A). For balanced accuracy the
distribution of fairness and performance values was slightly more
complex, exhibiting a slight fairness-performance trade-off
(Figure~\ref{fig-performance-fairness-acc} B).

\hypertarget{tbl-varimp}{}
\begin{table}
\caption{\label{tbl-varimp}The 10 most important decisions or decision interactions and their
relative importance. }\tabularnewline

\centering
\begin{tabular}{llrr}
\toprule
Effect Type & Decision / Interaction of Decisions & Importance & Std. Deviation\\
\midrule
main & $\mathit{Stratify Split}$ & 0.375 & 0.001\\
2-way int. & $\mathit{Cutoff} \times \mathit{Stratify Split}$ & 0.313 & 0.000\\
main & $\mathit{Cutoff}$ & 0.081 & 0.000\\
4-way int. & $\mathit{Cutoff} \times \mathit{Exclude Features} \times \mathit{Model} \times \mathit{Stratify Split}$ & 0.008 & 0.000\\
3-way int. & $\mathit{Cutoff} \times \mathit{Model} \times \mathit{Stratify Split}$ & 0.007 & 0.000\\
\addlinespace
3-way int. & $\mathit{Cutoff} \times \mathit{Model} \times \mathit{Preprocess Income}$ & 0.007 & 0.000\\
2-way int. & $\mathit{Model} \times \mathit{Preprocess Income}$ & 0.007 & 0.000\\
2-way int. & $\mathit{Exclude Features} \times \mathit{Model}$ & 0.006 & 0.000\\
3-way int. & $\mathit{Model} \times \mathit{Preprocess Income} \times \mathit{Scale}$ & 0.006 & 0.000\\
2-way int. & $\mathit{Cutoff} \times \mathit{Preprocess Income}$ & 0.005 & 0.000\\
\bottomrule
\end{tabular}
\end{table}

\hypertarget{importance-of-decisions}{%
\subsubsection{Importance of Decisions}\label{importance-of-decisions}}

We conducted a FANOVA \citep{hooker2007} as described in
\citet{hutter2014} to assess the importance of decisions on the fairness
metric. This analysis decomposes the overall variance of the fairness
metric into the fractions which are explained by each decision. These
variance decompositions are used to assess the relative importance of
decisions. Moreover, the FANOVA also allows computing explained variance
for interactions of decisions. This is highly useful, as the overall
interaction space between decisions is quite large with \(511\) possible
(interaction and main) effects.

Using the resulting importance values from the FANOVA, one can see which
decisions are associated with a high variation in fairness scores,
whether it be by themselves or in conjunction with others. This allows
assessing the most consequential decisions on a one-by-one case. Table~\ref{tbl-varimp} contains a ranked list of the most important decisions
and decision interactions in our case study alongside their respective
importance.

As can be seen in Table~\ref{tbl-varimp}, the most important decision is how the stratification of
the train-test split is performed. Moreover, the interaction of the
chosen cutoff value with the stratification strategy is highly
important, accounting for more than \(30\%\) of the variance in the
fairness metric. It also becomes apparent that especially the
\emph{interactions} of decisions are relevant here, with all decisions
among the top 10 except the stratification and cutoff being interactions
rather than sole decisions.

\begin{figure}
{\centering \includegraphics[width=0.95\textwidth]{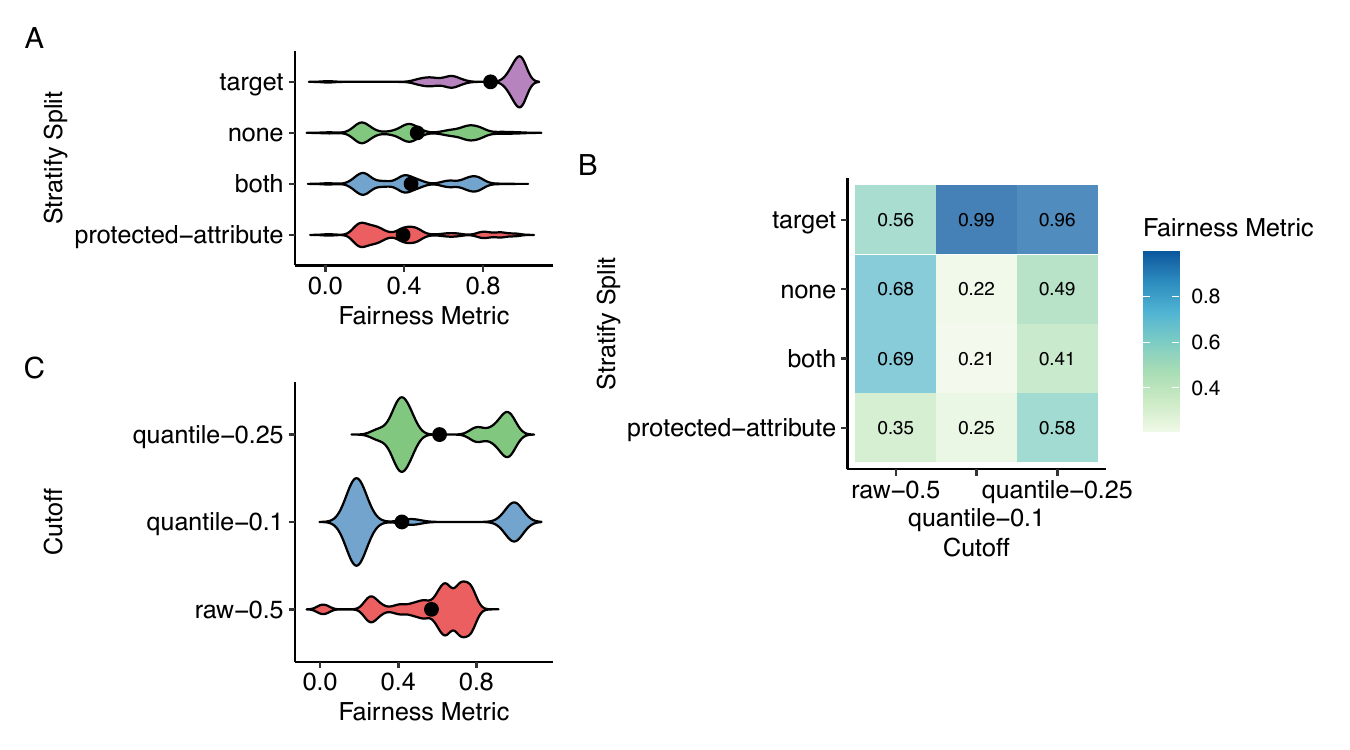}
}
\caption{\label{fig-top-decisions}\textbf{The influence of decisions on the fairness metric can only be understood when examining interactions on top of individual decisions.} Visualization of the fairness metric depending on the three most important decision / decision combinations (from A - C by importance) and their respective options.}

\end{figure}

We analyzed the three most important decisions or decision-interactions to further illustrate the methodology and how one would explore the results of the analysis. The results also highlight why one should investigate the decisions in a detailed manner and not just pick the most-fair and highest-performing universe's model. The decisions \emph{Stratify Split}, \emph{Cutoff} and their interaction account for all three of the most important decisions. When examining the decision separately, it can be seen how stratifying by the target variable leads to noticeably lower fairness scores (Figure~\ref{fig-top-decisions} A, most important) and how the raw cutoff value of \(0.5\) is suddenly not leading to the best fairness scores anymore (Figure~\ref{fig-top-decisions} B, third most important). The effects of both variables become most clear, however, when examining their interaction, which was identified as explaining almost as much variance as the most important decision. While using a cutoff value corresponding to the top \(10\%\) quantile leads to the least fair model when stratifying by the target variable it surprisingly leads to the models with the best average fairness metric when using any other stratification strategy (Figure~\ref{fig-top-decisions} C, second most important).

As variation in random train-test splits can affect fairness and performance of machine learning models \citep{cooper2024arbitrariness,friedler2019comparative}, we repeated the complete multiverse analysis five times with different random seeds, achieving highly similar results regarding both the overall variation of the fairness metric (Figure~\ref{fig-var-replications}) and the relative importance of decisions (Figure~\ref{fig-cor-replications}).

\hypertarget{scaling-a-design-multiverse-analysis-for-algorithmic-fairness}{%
\subsubsection{Scaling the Analysis}\label{scaling-a-design-multiverse-analysis-for-algorithmic-fairness}}

Conducting a multiverse analysis can be computationally expensive.
Especially if the multiverse is particularly large or computational resources
are limited, it may not be possible to explore the complete grid of universes. To assess the feasibility of running the multiverse analysis on a
smaller subset of the grid, we also conducted the FANOVAs on different
subsamples of the collected \emph{multiverse} dataset. Specifically, we
ran the analysis on random subsets of \(1\%\), \(5\%\), \(10\%\) and
\(20\%\) of the data and calculated the correlation of variance
decomposition or importance values with the FANOVA estimated on the full
multiverse dataset. The estimates of variance decomposition are highly
skewed, with a few highly important decisions and a very larger number
of very low-importance decisions. We therefore calculated both, the
Pearson correlation which is more sensitive to correlations of the more
important decisions and the Spearman rank-correlation which is also
sensitive to decisions with low importance estimates. To assess the
consistency of this approach we computed the FANOVA on each subsample 50
times and calculated the correlation with the results from the full
\emph{multiverse} dataset every time.

When calculating the Pearson correlation, the resulting mean correlation
coefficient ranged from \(\bar{r}_{1\%} =0.996\) (\(SD = 0.003\)) at
\(1\%\) to \(\bar{r}_{20\%} \geq 0.999\) (\(SD = 0\)) at \(20\%\).
Spearman rank-correlations were also high, but lower than the Pearson
correlation coefficients and more inconsistent
(Figure~\ref{fig-correlations}), which indicates that using sparse data
to estimate the importance of decisions works well for important decisions and less-so to identify nuances between less-important decisions. The resulting Spearman rank-correlation mean
coefficients ranged from \(\bar{\rho}_{1\%} =0.529\) (\(SD = 0.031\)) at
\(1\%\) to \(\bar{\rho}_{20\%} =0.937\) (\(SD = 0.007\)) at \(20\%\).

\hypertarget{study-2-evaluation-1}{%
\subsection{Study 2: Evaluation}\label{study-2-evaluation-1}}

By combining the different evaluation decisions we end up with
\(N = 28\) possible evaluation strategies for any given model. We
computed each of these for each of the universes from Study 1. This lead
to a total of \(N = 1,720,320\) values of the fairness metric with a
mean value of \(M = 0.339\). Similar to Study 1, these fairness values
exhibited a high degree of variation. However, variation stayed high,
even when examining values for the \emph{exact same model}. We observe a
full spread of the fairness metric from \(0\) to \(1\) (\(\Delta = 1\))
for 5.80\% of the models, only by varying their evaluation. Alarmingly,
we observe a spread of at least \(\Delta \geq 0.9\) on the fairness
metric for 94.51\% of models. In the following we examine variation due
to evaluation decisions for a single model in more detail.

\begin{figure}

{\centering \includegraphics[width=1.0\textwidth]{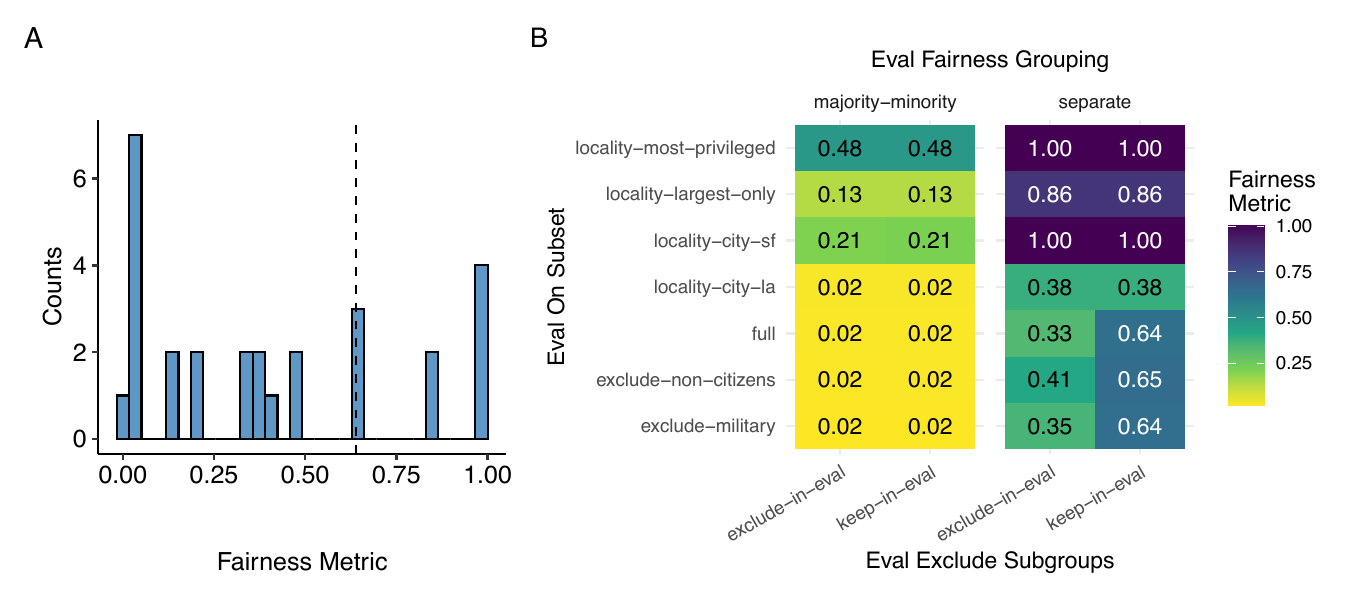}

}

\caption{\label{fig-eval-med}\textbf{The fairness metric of the exact same model can be significantly altered by varying its evaluation strategy alone (A) and especially the interaction of
different evaluation decisions leads to changes in the fairness metric (B).} Overall distribution (A) and raw values (B) of fairness metric (equalized odds difference) for a single model over different decisions regarding its evaluation. The dashed line in A corresponds to the evaluation strategy used in Study 1\footref{fn1eval}. Both plots display scores for a model showing median variation, to see the
same figure for the model with high variation see
Figure~\ref{fig-eval-max} in the Appendix. An \href{https://reliable-ai.github.io/fairml-multiverse/}{interactive version} of A is available, allowing examination of the distribution for any model in the multiverse analysis.}

\end{figure}

We examined the variation of two individual models in more detail to illustrate the impact of evaluation decisions on algorithmic fairness
for a single model. We chose to illustrate our point with one model exhibiting a median degree of variance based on evaluation decisions and one exhibiting a high degree. Neither model
resulted from a particularly extreme combination of options.\footnote{The
  options for the model with median variance are: Cutoff = raw-0.5,
  Encode Categorical = ordinal, Exclude Features = race, Exclude
  Subgroups = drop-smallest-2, Model = rf, Preprocess Age = quantiles-4,
  Preprocess Income = bins-10000, Scale = scale, Stratify Split = none. The
  options for the model with high variance are: Cutoff = quantile-0.1,
  Encode Categorical = one-hot, Exclude Features = race, Exclude
  Subgroups = drop-other, Model = rf, Preprocess Age = quantiles-4,
  Preprocess Income = none, Scale = scale, Stratify Split = none.}

The overall distribution of the fairness metric alongside a detailed breakdown by decisions can be seen in Figure~\ref{fig-eval-med} for the model with median variation and Figure~\ref{fig-eval-max} for the model with high variation. Under the evaluation strategy used in Study 1, the chosen model with high variance would be considered highly unfair with a metric of \(m_{EqOdds} = 1.000\) and the model with median variance slightly fairer with \(m_{EqOdds} = 0.638\). However, as can be seen in Figure~\ref{fig-eval-med}, there exist ample opportunities to tweak the evaluation strategy to achieve significantly better scores on the fairness metric. Indeed, both models can achieve a perfect score of 0 on the fairness metric, only by varying how they are evaluated. Given that the models stay exactly the same, we consider this practice ``fairness hacking''.

An overview of how evaluation decisions affect the fairness metric across the complete multiverse can be seen in Figure~\ref{fig-eval-decisions}, illustrating how e.g. the fairness grouping can consistently mask disparate treatment of minority groups. 

\hypertarget{discussion}{%
\section{Discussion}\label{discussion}}

We demonstrate how multiverse analysis for algorithmic fairness provides a useful new method for evaluating the robustness of machine learning and ADM systems with respect to decisions along the modeling pipeline and their implications for algorithmic fairness. We highlight the importance of making decisions during model design and evaluation explicitly rather than implicitly.

By applying this new methodology in a use case of predicting public health care coverage, we demonstrate the feasibility of this approach as well as how fairness metrics can be manipulated through evaluation strategies. We further show which decisions during model design affect fairness the most: Surprisingly, we see that the stratification strategy used for the train-test split has strong effects on the fairness metric. We also observe that the cutoff value used for making final decisions is important, a decision often implemented post-hoc after model deployment without consideration of fairness.

When interpreting the results from a multiverse analysis for algorithmic fairness, one should evaluate results with care and strictly avoid merely selecting the combination of decisions with the best fairness metric. Results should be seen as an indication of how susceptible the fairness of a model is to design decisions and which decisions warrant closer examination. Relative scores of decision importance should always be interpreted in light of the overall degree of observed variation. Results from the analysis can also be used to guide the search of new options for the most important decisions. Final choices regarding the design of the system should be made using a combination of empirical results from the multiverse analysis and practical as well as ethical considerations within the context of the use case. The main goal of a multiverse analysis for algorithmic fairness is to facilitate making educated and explicit decisions. We recommend including complete results from the analysis alongside the final system.

As we explored only a single use case, we do not make any generalizable claims regarding the importance of any particular decisions, beyond the fact that these decisions \emph{can} matter and are worth investigating. Another limitation of this case study is that we only examined nine design and three evaluation decisions, with many plausible alternative decisions which could have been examined in their place or additionally. As there is an infinite space of decisions one may consider, we decided to draw the line at these decisions for illustrative purposes. A successful adoption of multiverse analysis for algorithmic fairness in different use cases and reporting of results could help identify a more exhaustive list of the most important decisions across contexts. Potential concerns regarding the computational cost of conducting a multiverse analysis for algorithmic fairness are valid, but can be addressed as we demonstrate that important decisions are robustly detected even when exploring only \(1\%\) of the full \emph{multiverse}.

There are varying degrees of conducting a multiverse analysis of algorithmic fairness, each providing unique value and requiring different amounts of computation: We believe there is already significant value in (1) merely thinking about (implicit) decisions taken during system design and the consideration of potential alternatives, (2) performing a multiverse analysis of a fixed model with different evaluation strategies as a computationally inexpensive option to provide more robust evaluations and combat fairness hacking, (3) conducting a partial multiverse analysis of a subset of the full multiverse (e.g. 1\%) and (4) an analysis of the full multiverse as the most thorough option.

We encourage the use of the method during the design of future machine learning or ADM systems and provide an overview of the most important areas of decisions to guide analysts when adapting multiverse analysis for algorithmic fairness in their own context. We further provide a non-exhaustive list of exemplary decisions to serve as inspiration to identify potentially relevant decisions and source code that makes adoption to different use cases easy. We posit that results from a multiverse analysis for algorithmic fairness can critically inform discussions between developers and stakeholders and advise joint reflections on the ultimate design of ADM systems. We further advocate for the use of multiverse analysis in fairness evaluations to understand the distribution of fairness scores that can be evoked by the same model under different evaluation scenarios and to reduce the risk of potential fairness hacking by transparently reporting the entirety of results.

\section*{Research Ethics and Social Impact}

\subsection*{Ethics Statement}

Our selection of preprocessing and evaluation decisions builds on common practices observed in machine learning publications. While some of these practices such as excluding minority groups in preprocessing and evaluation are highly questionable and should not be normalized, we decided to include them in our case study to highlight their fairness implications and stimulate critical reflection. We further decided that criticism of individual manuscripts which implement such practices would not add much utility to our work, while potentially leading to (limited) negative consequences for their authors. Therefore, we present the implications of such data practices without singling out individual manuscripts.

\subsection*{Positionality Statement}

All authors are affiliated with organizations from Western, Educated, Industrialized, Rich, and Democratic (WEIRD) countries, in line with a common pattern in the fair ML research community \citep{septiandri2023weird}. This background inherently influenced the research practice of this study, including the case study and data that was chosen, which ultimately predetermined the design and evaluation decisions we focused on. We posit, however, that the proposed methodology can be applied in a wide range of contexts, tasks, and with various different data modalities and protected attributes.

\subsection*{Adverse Impact Statement}

We condemn potential misuses of our proposed method that contrast its objective of promoting transparency and reliability in machine learning practice and identified the following potential adverse impacts and misconceptions.

\begin{itemize}
    \item We do not interpret fairness as an optimization problem. A multiverse analysis allows to understand the \textit{variation of fairness scores} as a result of design decisions that researchers and developers might not have related to fairness in standard modeling practice and although fairness scores can imply real fairness they are only an indicator and not proof of fairness. While its results can inform discussions on sensible design decisions, the social impacts of an ADM system can only be understood by considering its specific implementation context and the interactions with the social environment in which it is placed. 
    \item A multiverse analysis critically depends on the careful identification of \textit{relevant design decisions}. While the decisions we examined in our case study may serve as a starting point, they do not present an exhaustive list by any means. Specifying a multiverse analysis requires researchers to carefully reflect on the data practices, processing and modeling decisions, embedded in their respective application context. 
    \item A multiverse analysis should not be used to search for the evaluation strategy which displays the best fairness score. On the contrary, it presents a tool whose usage can be requested by stakeholders to instead \textit{prevent selective reporting} and promote transparency by presenting the distribution of fairness scores across multiple evaluation schemes. It re-centers the discussion on how and for whom fairness metrics are computed, and acknowledges the susceptibility and instability of metrics to (small) changes in the evaluation protocol.
\end{itemize}

%% acknowledgments 
\begin{acks}

This work is supported by the DAAD programme Konrad Zuse Schools of Excellence in Artificial Intelligence, sponsored by the Federal Ministry of Education and Research, the Munich Center for Machine Learning and the Federal Statistical Office of Germany.

\end{acks}

\bibliographystyle{ACM-Reference-Format}
\bibliography{bibliography.bib}


\begin{thebibliography}{65}

%%% the Web address.

\ifx \showCODEN    \undefined \def \showCODEN     #1{\unskip}     \fi
\ifx \showDOI      \undefined \def \showDOI       #1{#1}\fi
\ifx \showISBNx    \undefined \def \showISBNx     #1{\unskip}     \fi
\ifx \showISBNxiii \undefined \def \showISBNxiii  #1{\unskip}     \fi
\ifx \showISSN     \undefined \def \showISSN      #1{\unskip}     \fi
\ifx \showLCCN     \undefined \def \showLCCN      #1{\unskip}     \fi
\ifx \shownote     \undefined \def \shownote      #1{#1}          \fi
\ifx \showarticletitle \undefined \def \showarticletitle #1{#1}   \fi
\ifx \showURL      \undefined \def \showURL       {\relax}        \fi
% invisible to TeX
\providecommand\bibfield[2]{#2}
\providecommand\bibinfo[2]{#2}
\providecommand\natexlab[1]{#1}
\providecommand\showeprint[2][]{arXiv:#2}

\bibitem[Agarwal et~al\mbox{.}(2018)]%
        {agarwal}
\bibfield{author}{\bibinfo{person}{Alekh Agarwal}, \bibinfo{person}{Alina Beygelzimer}, \bibinfo{person}{Miroslav {Dudík}}, \bibinfo{person}{John Langford}, {and} \bibinfo{person}{Hanna Wallach}.} \bibinfo{year}{2018}\natexlab{}.
\newblock \showarticletitle{A Reductions Approach to Fair Classification}.
\newblock  (\bibinfo{year}{2018}).
\newblock


\bibitem[Agrawal et~al\mbox{.}(2021)]%
        {agrawal}
\bibfield{author}{\bibinfo{person}{Ashrya Agrawal}, \bibinfo{person}{Florian Pfisterer}, \bibinfo{person}{Bernd Bischl}, \bibinfo{person}{Francois Buet-Golfouse}, \bibinfo{person}{Srijan Sood}, \bibinfo{person}{Jiahao Chen}, \bibinfo{person}{Sameena Shah}, {and} \bibinfo{person}{Sebastian Vollmer}.} \bibinfo{year}{2021}\natexlab{}.
\newblock \showarticletitle{Debiasing classifiers: is reality at variance with expectation?}
\newblock  (\bibinfo{year}{2021}).
\newblock
\urldef\tempurl%
\url{https://doi.org/10.48550/arXiv.2011.02407}
\showDOI{\tempurl}


\bibitem[A{\"\i}vodji et~al\mbox{.}(2019)]%
        {aivodji2019fairwashing}
\bibfield{author}{\bibinfo{person}{Ulrich A{\"\i}vodji}, \bibinfo{person}{Hiromi Arai}, \bibinfo{person}{Olivier Fortineau}, \bibinfo{person}{S{\'e}bastien Gambs}, \bibinfo{person}{Satoshi Hara}, {and} \bibinfo{person}{Alain Tapp}.} \bibinfo{year}{2019}\natexlab{}.
\newblock \showarticletitle{Fairwashing: the risk of rationalization}. In \bibinfo{booktitle}{\emph{International Conference on Machine Learning}}. PMLR, \bibinfo{pages}{161--170}.
\newblock


\bibitem[Allaire et~al\mbox{.}(2022)]%
        {allaire2022}
\bibfield{author}{\bibinfo{person}{J.J. Allaire}, \bibinfo{person}{Charles Teague}, \bibinfo{person}{Carlos Scheidegger}, \bibinfo{person}{Yihui Xie}, {and} \bibinfo{person}{Christophe Dervieux}.} \bibinfo{year}{2022}\natexlab{}.
\newblock \bibinfo{booktitle}{\emph{Quarto}}.
\newblock
\urldef\tempurl%
\url{https://doi.org/10.5281/zenodo.5960048}
\showDOI{\tempurl}
\newblock
\shownote{DOI: 10.5281/zenodo.5960048}.


\bibitem[Angwin et~al\mbox{.}(2016)]%
        {angwin2016}
\bibfield{author}{\bibinfo{person}{Julia Angwin}, \bibinfo{person}{Jeff Larson}, \bibinfo{person}{Surya Mattu}, {and} \bibinfo{person}{Lauren Kirchner}.} \bibinfo{year}{2016}\natexlab{}.
\newblock \showarticletitle{Machine bias}.
\newblock \bibinfo{journal}{\emph{ProPublica}} (\bibinfo{date}{05} \bibinfo{year}{2016}), \bibinfo{pages}{254{\textendash}264}.
\newblock
\urldef\tempurl%
\url{https://www.propublica.org/article/machine-bias-risk-assessments-in-criminal-sentencing}
\showURL{%
\tempurl}


\bibitem[Bao et~al\mbox{.}(2022)]%
        {bao}
\bibfield{author}{\bibinfo{person}{Michelle Bao}, \bibinfo{person}{Angela Zhou}, \bibinfo{person}{Samantha Zottola}, \bibinfo{person}{Brian Brubach}, \bibinfo{person}{Sarah Desmarais}, \bibinfo{person}{Aaron Horowitz}, \bibinfo{person}{Kristian Lum}, {and} \bibinfo{person}{Suresh Venkatasubramanian}.} \bibinfo{year}{2022}\natexlab{}.
\newblock \showarticletitle{It's COMPASlicated: The Messy Relationship between RAI Datasets and Algorithmic Fairness Benchmarks}.
\newblock  (\bibinfo{year}{2022}).
\newblock
\urldef\tempurl%
\url{https://doi.org/10.48550/arXiv.2106.05498}
\showDOI{\tempurl}


\bibitem[Barocas et~al\mbox{.}(2023)]%
        {barocas2023classification}
\bibfield{author}{\bibinfo{person}{Solon Barocas}, \bibinfo{person}{Moritz Hardt}, {and} \bibinfo{person}{Arvind Narayanan}.} \bibinfo{year}{2023}\natexlab{}.
\newblock \showarticletitle{Classification - {{No}} Fairness through Unawareness}.
\newblock In \bibinfo{booktitle}{\emph{Fairness and Machine Learning: Limitations and Opportunities}}. \bibinfo{publisher}{{The MIT Press}}, \bibinfo{address}{{Cambridge, Massachusetts}}.
\newblock
\showISBNx{978-0-262-04861-3}


\bibitem[Bell et~al\mbox{.}(2022)]%
        {bell}
\bibfield{author}{\bibinfo{person}{Samuel~J. Bell}, \bibinfo{person}{Onno~P. Kampman}, \bibinfo{person}{Jesse Dodge}, {and} \bibinfo{person}{Neil~D. Lawrence}.} \bibinfo{year}{2022}\natexlab{}.
\newblock \showarticletitle{Modeling the Machine Learning Multiverse}.
\newblock  (\bibinfo{year}{2022}).
\newblock
\urldef\tempurl%
\url{https://doi.org/10.48550/arXiv.2206.05985}
\showDOI{\tempurl}


\bibitem[Bischl et~al\mbox{.}(2023)]%
        {bischl2023}
\bibfield{author}{\bibinfo{person}{Bernd Bischl}, \bibinfo{person}{Martin Binder}, \bibinfo{person}{Michel Lang}, \bibinfo{person}{Tobias Pielok}, \bibinfo{person}{Jakob Richter}, \bibinfo{person}{Stefan Coors}, \bibinfo{person}{Janek Thomas}, \bibinfo{person}{Theresa Ullmann}, \bibinfo{person}{Marc Becker}, \bibinfo{person}{{Anne{-}Laure} Boulesteix}, \bibinfo{person}{Difan Deng}, {and} \bibinfo{person}{Marius Lindauer}.} \bibinfo{year}{2023}\natexlab{}.
\newblock \showarticletitle{Hyperparameter optimization: Foundations, algorithms, best practices, and open challenges}.
\newblock \bibinfo{journal}{\emph{WIREs Data Mining and Knowledge Discovery}} \bibinfo{volume}{13}, \bibinfo{number}{2} (\bibinfo{date}{03} \bibinfo{year}{2023}).
\newblock
\urldef\tempurl%
\url{https://doi.org/10.1002/widm.1484}
\showDOI{\tempurl}


\bibitem[Black et~al\mbox{.}(2022)]%
        {black2022}
\bibfield{author}{\bibinfo{person}{Emily Black}, \bibinfo{person}{Manish Raghavan}, {and} \bibinfo{person}{Solon Barocas}.} \bibinfo{year}{2022}\natexlab{}.
\newblock \showarticletitle{Model Multiplicity: Opportunities, Concerns, and Solutions}.
\newblock  (\bibinfo{year}{2022}).
\newblock


\bibitem[Breznau et~al\mbox{.}(2022)]%
        {breznau2022}
\bibfield{author}{\bibinfo{person}{Nate Breznau}, \bibinfo{person}{Eike~Mark Rinke}, \bibinfo{person}{Alexander Wuttke}, \bibinfo{person}{Hung H.~V. Nguyen}, \bibinfo{person}{Muna Adem}, \bibinfo{person}{Jule Adriaans}, \bibinfo{person}{Amalia Alvarez-Benjumea}, \bibinfo{person}{Henrik~K. Andersen}, \bibinfo{person}{Daniel Auer}, \bibinfo{person}{Flavio Azevedo}, \bibinfo{person}{Oke Bahnsen}, \bibinfo{person}{Dave Balzer}, \bibinfo{person}{Gerrit Bauer}, \bibinfo{person}{Paul~C. Bauer}, \bibinfo{person}{Markus Baumann}, \bibinfo{person}{Sharon Baute}, \bibinfo{person}{Verena Benoit}, \bibinfo{person}{Julian Bernauer}, \bibinfo{person}{Carl Berning}, \bibinfo{person}{Anna Berthold}, \bibinfo{person}{Felix~S. Bethke}, \bibinfo{person}{Thomas Biegert}, \bibinfo{person}{Katharina Blinzler}, \bibinfo{person}{Johannes~N. Blumenberg}, \bibinfo{person}{Licia Bobzien}, \bibinfo{person}{Andrea Bohman}, \bibinfo{person}{Thijs Bol}, \bibinfo{person}{Amie Bostic}, \bibinfo{person}{Zuzanna Brzozowska},
  \bibinfo{person}{Katharina Burgdorf}, \bibinfo{person}{Kaspar Burger}, \bibinfo{person}{Kathrin~B. Busch}, \bibinfo{person}{Juan Carlos-Castillo}, \bibinfo{person}{Nathan Chan}, \bibinfo{person}{Pablo Christmann}, \bibinfo{person}{Roxanne Connelly}, \bibinfo{person}{Christian~S. Czymara}, \bibinfo{person}{Elena Damian}, \bibinfo{person}{Alejandro Ecker}, \bibinfo{person}{Achim Edelmann}, \bibinfo{person}{Maureen~A. Eger}, \bibinfo{person}{Simon Ellerbrock}, \bibinfo{person}{Anna Forke}, \bibinfo{person}{Andrea Forster}, \bibinfo{person}{Chris Gaasendam}, \bibinfo{person}{Konstantin Gavras}, \bibinfo{person}{Vernon Gayle}, \bibinfo{person}{Theresa Gessler}, \bibinfo{person}{Timo Gnambs}, \bibinfo{person}{{Amélie} Godefroidt}, \bibinfo{person}{Max {Grömping}}, \bibinfo{person}{Martin {Groß}}, \bibinfo{person}{Stefan Gruber}, \bibinfo{person}{Tobias Gummer}, \bibinfo{person}{Andreas Hadjar}, \bibinfo{person}{Jan~Paul Heisig}, \bibinfo{person}{Sebastian Hellmeier}, \bibinfo{person}{Stefanie Heyne},
  \bibinfo{person}{Magdalena Hirsch}, \bibinfo{person}{Mikael Hjerm}, \bibinfo{person}{Oshrat Hochman}, \bibinfo{person}{Andreas {Hövermann}}, \bibinfo{person}{Sophia Hunger}, \bibinfo{person}{Christian Hunkler}, \bibinfo{person}{Nora Huth}, \bibinfo{person}{{Zsófia S.} {Ignácz}}, \bibinfo{person}{Laura Jacobs}, \bibinfo{person}{Jannes Jacobsen}, \bibinfo{person}{Bastian Jaeger}, \bibinfo{person}{Sebastian Jungkunz}, \bibinfo{person}{Nils Jungmann}, \bibinfo{person}{Mathias Kauff}, \bibinfo{person}{Manuel Kleinert}, \bibinfo{person}{Julia Klinger}, \bibinfo{person}{Jan-Philipp Kolb}, \bibinfo{person}{Marta {Ko{\l}czy{\'{n}}ska}}, \bibinfo{person}{John Kuk}, \bibinfo{person}{Katharina {Kunißen}}, \bibinfo{person}{Dafina Kurti~Sinatra}, \bibinfo{person}{Alexander Langenkamp}, \bibinfo{person}{Philipp~M. Lersch}, \bibinfo{person}{Lea-Maria {Löbel}}, \bibinfo{person}{Philipp Lutscher}, \bibinfo{person}{Matthias Mader}, \bibinfo{person}{Joan~E. Madia}, \bibinfo{person}{Natalia Malancu}, \bibinfo{person}{Luis
  Maldonado}, \bibinfo{person}{Helge Marahrens}, \bibinfo{person}{Nicole Martin}, \bibinfo{person}{Paul Martinez}, \bibinfo{person}{Jochen Mayerl}, \bibinfo{person}{Oscar~J. Mayorga}, \bibinfo{person}{Patricia McManus}, \bibinfo{person}{Kyle McWagner}, \bibinfo{person}{Cecil Meeusen}, \bibinfo{person}{Daniel Meierrieks}, \bibinfo{person}{Jonathan Mellon}, \bibinfo{person}{Friedolin Merhout}, \bibinfo{person}{Samuel Merk}, \bibinfo{person}{Daniel Meyer}, \bibinfo{person}{Leticia Micheli}, \bibinfo{person}{Jonathan Mijs}, \bibinfo{person}{{Cristóbal} Moya}, \bibinfo{person}{Marcel Neunhoeffer}, \bibinfo{person}{Daniel {Nüst}}, \bibinfo{person}{Olav {Nygård}}, \bibinfo{person}{Fabian Ochsenfeld}, \bibinfo{person}{Gunnar Otte}, \bibinfo{person}{Anna~O. Pechenkina}, \bibinfo{person}{Christopher Prosser}, \bibinfo{person}{Louis Raes}, \bibinfo{person}{Kevin Ralston}, \bibinfo{person}{Miguel~R. Ramos}, \bibinfo{person}{Arne Roets}, \bibinfo{person}{Jonathan Rogers}, \bibinfo{person}{Guido Ropers},
  \bibinfo{person}{Robin Samuel}, \bibinfo{person}{Gregor Sand}, \bibinfo{person}{Ariela Schachter}, \bibinfo{person}{Merlin Schaeffer}, \bibinfo{person}{David Schieferdecker}, \bibinfo{person}{Elmar Schlueter}, \bibinfo{person}{Regine Schmidt}, \bibinfo{person}{Katja~M. Schmidt}, \bibinfo{person}{Alexander Schmidt-Catran}, \bibinfo{person}{Claudia Schmiedeberg}, \bibinfo{person}{{Jürgen} Schneider}, \bibinfo{person}{Martijn Schoonvelde}, \bibinfo{person}{Julia Schulte-Cloos}, \bibinfo{person}{Sandy Schumann}, \bibinfo{person}{Reinhard Schunck}, \bibinfo{person}{{Jürgen} Schupp}, \bibinfo{person}{Julian Seuring}, \bibinfo{person}{Henning Silber}, \bibinfo{person}{Willem Sleegers}, \bibinfo{person}{Nico Sonntag}, \bibinfo{person}{Alexander Staudt}, \bibinfo{person}{Nadia Steiber}, \bibinfo{person}{Nils Steiner}, \bibinfo{person}{Sebastian Sternberg}, \bibinfo{person}{Dieter Stiers}, \bibinfo{person}{Dragana Stojmenovska}, \bibinfo{person}{Nora Storz}, \bibinfo{person}{Erich Striessnig},
  \bibinfo{person}{Anne-Kathrin Stroppe}, \bibinfo{person}{Janna Teltemann}, \bibinfo{person}{Andrey Tibajev}, \bibinfo{person}{Brian Tung}, \bibinfo{person}{Giacomo Vagni}, \bibinfo{person}{Jasper Van~Assche}, \bibinfo{person}{Meta van~der Linden}, \bibinfo{person}{Jolanda van~der Noll}, \bibinfo{person}{Arno Van~Hootegem}, \bibinfo{person}{Stefan Vogtenhuber}, \bibinfo{person}{Bogdan Voicu}, \bibinfo{person}{Fieke Wagemans}, \bibinfo{person}{Nadja Wehl}, \bibinfo{person}{Hannah Werner}, \bibinfo{person}{Brenton~M. Wiernik}, \bibinfo{person}{Fabian Winter}, \bibinfo{person}{Christof Wolf}, \bibinfo{person}{Yuki Yamada}, \bibinfo{person}{Nan Zhang}, \bibinfo{person}{Conrad Ziller}, \bibinfo{person}{Stefan Zins}, {and} \bibinfo{person}{Tomasz {{\.{Z}}ó{\l}tak}}.} \bibinfo{year}{2022}\natexlab{}.
\newblock \showarticletitle{Observing many researchers using the same data and hypothesis reveals a hidden universe of uncertainty}.
\newblock \bibinfo{journal}{\emph{Proceedings of the National Academy of Sciences}} \bibinfo{volume}{119}, \bibinfo{number}{44} (\bibinfo{date}{11} \bibinfo{year}{2022}), \bibinfo{pages}{e2203150119}.
\newblock
\urldef\tempurl%
\url{https://doi.org/10.1073/pnas.2203150119}
\showDOI{\tempurl}
\newblock
\shownote{Publisher: Proceedings of the National Academy of Sciences}.


\bibitem[Bureau(2021a)]%
        {uscensusbureau}
\bibfield{author}{\bibinfo{person}{US~Census Bureau}.} \bibinfo{year}{2021}\natexlab{a}.
\newblock \bibinfo{title}{ACS Health Insurance Coverage Recoding Programming Code}.
\newblock
\newblock
\urldef\tempurl%
\url{https://www.census.gov/topics/health/health-insurance/guidance/programming-code/acs-recoding.html}
\showURL{%
\tempurl}
\newblock
\shownote{Section: Government}.


\bibitem[Bureau(2021b)]%
        {us2021understanding}
\bibfield{author}{\bibinfo{person}{US~Census Bureau}.} \bibinfo{year}{2021}\natexlab{b}.
\newblock \bibinfo{title}{Understanding and using the American Community Survey public use microdata sample files: What data users need to know}.
\newblock
\newblock


\bibitem[Caton et~al\mbox{.}(2022)]%
        {caton2022}
\bibfield{author}{\bibinfo{person}{Simon Caton}, \bibinfo{person}{Saiteja Malisetty}, {and} \bibinfo{person}{Christian Haas}.} \bibinfo{year}{2022}\natexlab{}.
\newblock \showarticletitle{Impact of Imputation Strategies on Fairness in Machine Learning}.
\newblock \bibinfo{journal}{\emph{Journal of Artificial Intelligence Research}}  \bibinfo{volume}{74} (\bibinfo{date}{09} \bibinfo{year}{2022}).
\newblock
\urldef\tempurl%
\url{https://doi.org/10.1613/jair.1.13197}
\showDOI{\tempurl}


\bibitem[contributors(2017)]%
        {contributors2017}
\bibfield{author}{\bibinfo{person}{nteract contributors}.} \bibinfo{year}{2017}\natexlab{}.
\newblock \bibinfo{booktitle}{\emph{papermill: Parametrize and run Jupyter and nteract Notebooks}}.
\newblock
\urldef\tempurl%
\url{https://github.com/nteract/papermill}
\showURL{%
\tempurl}


\bibitem[Cooper et~al\mbox{.}(2024)]%
        {cooper2024arbitrariness}
\bibfield{author}{\bibinfo{person}{A.~Feder Cooper}, \bibinfo{person}{Katherine Lee}, \bibinfo{person}{Madiha~Zahrah Choksi}, \bibinfo{person}{Solon Barocas}, \bibinfo{person}{Christopher De~Sa}, \bibinfo{person}{James Grimmelmann}, \bibinfo{person}{Jon Kleinberg}, \bibinfo{person}{Siddhartha Sen}, {and} \bibinfo{person}{Baobao Zhang}.} \bibinfo{year}{2024}\natexlab{}.
\newblock \showarticletitle{Arbitrariness and {{Social Prediction}}: {{The Confounding Role}} of {{Variance}} in {{Fair Classification}}}.
\newblock \bibinfo{journal}{\emph{Proceedings of the AAAI Conference on Artificial Intelligence}} \bibinfo{volume}{38}, \bibinfo{number}{20} (\bibinfo{date}{March} \bibinfo{year}{2024}), \bibinfo{pages}{22004--22012}.
\newblock
\showISSN{2374-3468, 2159-5399}
\urldef\tempurl%
\url{https://doi.org/10.1609/aaai.v38i20.30203}
\showDOI{\tempurl}


\bibitem[Cox(1958)]%
        {cox1958}
\bibfield{author}{\bibinfo{person}{David~R Cox}.} \bibinfo{year}{1958}\natexlab{}.
\newblock \showarticletitle{The regression analysis of binary sequences}.
\newblock \bibinfo{journal}{\emph{Journal of the Royal Statistical Society: Series B (Methodological)}} \bibinfo{volume}{20}, \bibinfo{number}{2} (\bibinfo{year}{1958}), \bibinfo{pages}{215{\textendash}232}.
\newblock
\newblock
\shownote{Publisher: Wiley Online Library}.


\bibitem[Ding et~al\mbox{.}(2021)]%
        {ding2021}
\bibfield{author}{\bibinfo{person}{Frances Ding}, \bibinfo{person}{Moritz Hardt}, \bibinfo{person}{John Miller}, {and} \bibinfo{person}{Ludwig Schmidt}.} \bibinfo{year}{2021}\natexlab{}.
\newblock \showarticletitle{Retiring Adult: New Datasets for Fair Machine Learning}.
\newblock  (\bibinfo{year}{2021}), \bibinfo{pages}{13}.
\newblock


\bibitem[Dooley et~al\mbox{.}(2024)]%
        {dooley2024rethinking}
\bibfield{author}{\bibinfo{person}{Samuel Dooley}, \bibinfo{person}{Rhea Sukthanker}, \bibinfo{person}{John Dickerson}, \bibinfo{person}{Colin White}, \bibinfo{person}{Frank Hutter}, {and} \bibinfo{person}{Micah Goldblum}.} \bibinfo{year}{2024}\natexlab{}.
\newblock \showarticletitle{Rethinking bias mitigation: Fairer architectures make for fairer face recognition}.
\newblock \bibinfo{journal}{\emph{Advances in Neural Information Processing Systems}}  \bibinfo{volume}{36} (\bibinfo{year}{2024}).
\newblock


\bibitem[Fabris et~al\mbox{.}(2022)]%
        {fabris2022}
\bibfield{author}{\bibinfo{person}{Alessandro Fabris}, \bibinfo{person}{Stefano Messina}, \bibinfo{person}{Gianmaria Silvello}, {and} \bibinfo{person}{Gian~Antonio Susto}.} \bibinfo{year}{2022}\natexlab{}.
\newblock \showarticletitle{Algorithmic fairness datasets: the story so far}.
\newblock \bibinfo{journal}{\emph{Data Mining and Knowledge Discovery}} (\bibinfo{date}{09} \bibinfo{year}{2022}).
\newblock
\urldef\tempurl%
\url{https://doi.org/10.1007/s10618-022-00854-z}
\showDOI{\tempurl}


\bibitem[Faliagka et~al\mbox{.}(2012)]%
        {faliagka2012}
\bibfield{author}{\bibinfo{person}{Evanthia Faliagka}, \bibinfo{person}{Kostas Ramantas}, {and} \bibinfo{person}{Giannis Tzimas}.} \bibinfo{year}{2012}\natexlab{}.
\newblock \showarticletitle{Application of Machine Learning Algorithms to an online Recruitment System}.
\newblock  (\bibinfo{year}{2012}).
\newblock


\bibitem[Feurer and Hutter(2019)]%
        {feurer2019}
\bibfield{author}{\bibinfo{person}{Matthias Feurer} {and} \bibinfo{person}{Frank Hutter}.} \bibinfo{year}{2019}\natexlab{}.
\newblock \bibinfo{booktitle}{\emph{Hyperparameter Optimization}}.
\newblock \bibinfo{publisher}{Springer International Publishing}, \bibinfo{address}{Cham}, \bibinfo{pages}{3--33}.
\newblock
\urldef\tempurl%
\url{https://doi.org/10.1007/978-3-030-05318-5_1}
\showDOI{\tempurl}
\newblock
\shownote{DOI: 10.1007/978-3-030-05318-5{\_}1}.


\bibitem[Friedler et~al\mbox{.}(2019)]%
        {friedler2019comparative}
\bibfield{author}{\bibinfo{person}{Sorelle~A Friedler}, \bibinfo{person}{Carlos Scheidegger}, \bibinfo{person}{Suresh Venkatasubramanian}, \bibinfo{person}{Sonam Choudhary}, \bibinfo{person}{Evan~P Hamilton}, {and} \bibinfo{person}{Derek Roth}.} \bibinfo{year}{2019}\natexlab{}.
\newblock \showarticletitle{A comparative study of fairness-enhancing interventions in machine learning}. In \bibinfo{booktitle}{\emph{Proceedings of the conference on fairness, accountability, and transparency}}. \bibinfo{pages}{329--338}.
\newblock


\bibitem[Friedman(2001)]%
        {friedman2001}
\bibfield{author}{\bibinfo{person}{Jerome~H. Friedman}.} \bibinfo{year}{2001}\natexlab{}.
\newblock \showarticletitle{Greedy function approximation: A gradient boosting machine.}
\newblock \bibinfo{journal}{\emph{The Annals of Statistics}} \bibinfo{volume}{29}, \bibinfo{number}{5} (\bibinfo{date}{10} \bibinfo{year}{2001}), \bibinfo{pages}{1189--1232}.
\newblock
\urldef\tempurl%
\url{https://doi.org/10.1214/aos/1013203451}
\showDOI{\tempurl}
\newblock
\shownote{Publisher: Institute of Mathematical Statistics}.


\bibitem[Gelman and Loken(2014)]%
        {gelman2014}
\bibfield{author}{\bibinfo{person}{Andrew Gelman} {and} \bibinfo{person}{Eric Loken}.} \bibinfo{year}{2014}\natexlab{}.
\newblock \showarticletitle{The Statistical Crisis in Science}.
\newblock \bibinfo{journal}{\emph{American Scientist}} \bibinfo{volume}{102}, \bibinfo{number}{6} (\bibinfo{year}{2014}), \bibinfo{pages}{460}.
\newblock
\newblock
\shownote{Publisher: Sigma XI-The Scientific Research Society}.


\bibitem[Hardt et~al\mbox{.}(2016)]%
        {hardt}
\bibfield{author}{\bibinfo{person}{Moritz Hardt}, \bibinfo{person}{Eric Price}, \bibinfo{person}{Eric Price}, {and} \bibinfo{person}{Nati Srebro}.} \bibinfo{year}{2016}\natexlab{}.
\newblock \showarticletitle{Equality of Opportunity in Supervised Learning}.
\newblock  (\bibinfo{year}{2016}).
\newblock


\bibitem[Henriques-Gomes(2023)]%
        {henriques-gomes2023}
\bibfield{author}{\bibinfo{person}{Luke Henriques-Gomes}.} \bibinfo{year}{2023}\natexlab{}.
\newblock \showarticletitle{Robodebt: five years of lies, mistakes and failures that caused a {\$}1.8bn scandal}.
\newblock \bibinfo{journal}{\emph{The Guardian}} (\bibinfo{date}{03} \bibinfo{year}{2023}).
\newblock
\urldef\tempurl%
\url{https://www.theguardian.com/australia-news/2023/mar/11/robodebt-five-years-of-lies-mistakes-and-failures-that-caused-a-18bn-scandal}
\showURL{%
\tempurl}


\bibitem[Ho(1995)]%
        {ho1995}
\bibfield{author}{\bibinfo{person}{Tin~Kam Ho}.} \bibinfo{year}{1995}\natexlab{}.
\newblock \showarticletitle{Random decision forests}, Vol.~\bibinfo{volume}{1}. \bibinfo{publisher}{IEEE}, \bibinfo{pages}{278{\textendash}282}.
\newblock


\bibitem[Hooker(2007)]%
        {hooker2007}
\bibfield{author}{\bibinfo{person}{Giles Hooker}.} \bibinfo{year}{2007}\natexlab{}.
\newblock \showarticletitle{Generalized Functional ANOVA Diagnostics for High-Dimensional Functions of Dependent Variables}.
\newblock \bibinfo{journal}{\emph{Journal of Computational and Graphical Statistics}} \bibinfo{volume}{16}, \bibinfo{number}{3} (\bibinfo{year}{2007}), \bibinfo{pages}{709--732}.
\newblock
\urldef\tempurl%
\url{https://www.jstor.org/stable/27594267}
\showURL{%
\tempurl}


\bibitem[Hutter et~al\mbox{.}(2014)]%
        {hutter2014}
\bibfield{author}{\bibinfo{person}{Frank Hutter}, \bibinfo{person}{Holger Hoos}, {and} \bibinfo{person}{Kevin Leyton-Brown}.} \bibinfo{year}{2014}\natexlab{}.
\newblock \showarticletitle{International Conference on Machine Learning}. \bibinfo{publisher}{PMLR}, \bibinfo{pages}{754--762}.
\newblock
\urldef\tempurl%
\url{https://proceedings.mlr.press/v32/hutter14.html}
\showURL{%
\tempurl}
\newblock
\shownote{ISSN: 1938-7228}.


\bibitem[International(2021)]%
        {amnestyinternational2021}
\bibfield{author}{\bibinfo{person}{Amnesty International}.} \bibinfo{year}{2021}\natexlab{}.
\newblock \bibinfo{booktitle}{\emph{Xenophobic Machines}}.
\newblock \bibinfo{type}{{T}echnical {R}eport}.
\newblock
\urldef\tempurl%
\url{https://www.amnesty.org/en/wp-content/uploads/2021/10/EUR3546862021ENGLISH.pdf}
\showURL{%
\tempurl}


\bibitem[Islam et~al\mbox{.}(2021)]%
        {islam2021}
\bibfield{author}{\bibinfo{person}{Rashidul Islam}, \bibinfo{person}{Shimei Pan}, {and} \bibinfo{person}{James~R. Foulds}.} \bibinfo{year}{2021}\natexlab{}.
\newblock \showarticletitle{AIES '21: AAAI/ACM Conference on AI, Ethics, and Society}. \bibinfo{publisher}{ACM}, \bibinfo{address}{Virtual Event USA}, \bibinfo{pages}{586--596}.
\newblock
\urldef\tempurl%
\url{https://doi.org/10.1145/3461702.3462614}
\showDOI{\tempurl}


\bibitem[Kassambara(2023)]%
        {kassambara2023}
\bibfield{author}{\bibinfo{person}{Alboukadel Kassambara}.} \bibinfo{year}{2023}\natexlab{}.
\newblock \bibinfo{booktitle}{\emph{ggpubr: 'ggplot2' Based Publication Ready Plots}}.
\newblock
\urldef\tempurl%
\url{https://CRAN.R-project.org/package=ggpubr}
\showURL{%
\tempurl}


\bibitem[Keisler-Starkey and Bunch(2022)]%
        {keisler-starkey2022}
\bibfield{author}{\bibinfo{person}{Katherine Keisler-Starkey} {and} \bibinfo{person}{Lisa~N Bunch}.} \bibinfo{year}{2022}\natexlab{}.
\newblock \bibinfo{booktitle}{\emph{Health Insurance Coverage in the United States: 2021 - Appendix Table C3}}.
\newblock \bibinfo{type}{{T}echnical {R}eport}.
\newblock
\urldef\tempurl%
\url{https://www.census.gov/content/dam/Census/library/publications/2022/demo/p60-278.pdf}
\showURL{%
\tempurl}


\bibitem[Kern et~al\mbox{.}(2021)]%
        {kern}
\bibfield{author}{\bibinfo{person}{Christoph Kern}, \bibinfo{person}{Ruben~L. Bach}, \bibinfo{person}{Hannah Mautner}, {and} \bibinfo{person}{Frauke Kreuter}.} \bibinfo{year}{2021}\natexlab{}.
\newblock \showarticletitle{Fairness in Algorithmic Profiling: A German Case Study}.
\newblock  (\bibinfo{year}{2021}).
\newblock
\urldef\tempurl%
\url{https://doi.org/10.48550/arXiv.2108.04134}
\showDOI{\tempurl}


\bibitem[Kohavi and Becker(1996)]%
        {kohavi1996}
\bibfield{author}{\bibinfo{person}{Ronny Kohavi} {and} \bibinfo{person}{Barry Becker}.} \bibinfo{year}{1996}\natexlab{}.
\newblock \showarticletitle{Adult data set}.
\newblock \bibinfo{journal}{\emph{UCI machine learning repository}}  \bibinfo{volume}{5} (\bibinfo{year}{1996}), \bibinfo{pages}{2093}.
\newblock


\bibitem[Kuhn and Johnson(2020)]%
        {kuhn2020}
\bibfield{author}{\bibinfo{person}{Max Kuhn} {and} \bibinfo{person}{Kjell Johnson}.} \bibinfo{year}{2020}\natexlab{}.
\newblock \bibinfo{booktitle}{\emph{Feature engineering and selection: a practical approach for predictive models}}.
\newblock \bibinfo{publisher}{CRC Press, Taylor \& Francis Group}, \bibinfo{address}{Boca Raton London New York}.
\newblock
\urldef\tempurl%
\url{www.feat.engineering}
\showURL{%
\tempurl}


\bibitem[Le~Quy et~al\mbox{.}(2022)]%
        {lequy2022}
\bibfield{author}{\bibinfo{person}{Tai Le~Quy}, \bibinfo{person}{Arjun Roy}, \bibinfo{person}{Vasileios Iosifidis}, \bibinfo{person}{Wenbin Zhang}, {and} \bibinfo{person}{Eirini Ntoutsi}.} \bibinfo{year}{2022}\natexlab{}.
\newblock \showarticletitle{A survey on datasets for fairness-aware machine learning}.
\newblock \bibinfo{journal}{\emph{WIREs Data Mining and Knowledge Discovery}} \bibinfo{volume}{12}, \bibinfo{number}{3} (\bibinfo{year}{2022}), \bibinfo{pages}{e1452}.
\newblock
\urldef\tempurl%
\url{https://doi.org/10.1002/widm.1452}
\showDOI{\tempurl}
\newblock
\shownote{{\_}eprint: https://onlinelibrary.wiley.com/doi/pdf/10.1002/widm.1452}.


\bibitem[Meding and Hagendorff(2024)]%
        {meding2024fairness}
\bibfield{author}{\bibinfo{person}{Kristof Meding} {and} \bibinfo{person}{Thilo Hagendorff}.} \bibinfo{year}{2024}\natexlab{}.
\newblock \showarticletitle{Fairness {{Hacking}}: {{The Malicious Practice}} of {{Shrouding Unfairness}} in {{Algorithms}}}.
\newblock \bibinfo{journal}{\emph{Philosophy \& Technology}} \bibinfo{volume}{37}, \bibinfo{number}{1} (\bibinfo{date}{Jan.} \bibinfo{year}{2024}), \bibinfo{pages}{4}.
\newblock
\showISSN{2210-5441}
\urldef\tempurl%
\url{https://doi.org/10.1007/s13347-023-00679-8}
\showDOI{\tempurl}


\bibitem[Mehrabi et~al\mbox{.}(2021)]%
        {mehrabi2021}
\bibfield{author}{\bibinfo{person}{Ninareh Mehrabi}, \bibinfo{person}{Fred Morstatter}, \bibinfo{person}{Nripsuta Saxena}, \bibinfo{person}{Kristina Lerman}, {and} \bibinfo{person}{Aram Galstyan}.} \bibinfo{year}{2021}\natexlab{}.
\newblock \showarticletitle{A Survey on Bias and Fairness in Machine Learning}.
\newblock \bibinfo{journal}{\emph{Comput. Surveys}} \bibinfo{volume}{54}, \bibinfo{number}{6} (\bibinfo{date}{07} \bibinfo{year}{2021}), \bibinfo{pages}{115:1{\textendash}115:35}.
\newblock
\urldef\tempurl%
\url{https://doi.org/10.1145/3457607}
\showDOI{\tempurl}


\bibitem[Meyer et~al\mbox{.}(2023)]%
        {meyer2023dataset}
\bibfield{author}{\bibinfo{person}{Anna~P. Meyer}, \bibinfo{person}{Aws Albarghouthi}, {and} \bibinfo{person}{Loris D'Antoni}.} \bibinfo{year}{2023}\natexlab{}.
\newblock \showarticletitle{The {{Dataset Multiplicity Problem}}: {{How Unreliable Data Impacts Predictions}}}. In \bibinfo{booktitle}{\emph{Proceedings of the 2023 {{ACM Conference}} on {{Fairness}}, {{Accountability}}, and {{Transparency}}}} \emph{(\bibinfo{series}{{{FAccT}} '23})}. \bibinfo{publisher}{Association for Computing Machinery}, \bibinfo{address}{New York, NY, USA}, \bibinfo{pages}{193--204}.
\newblock
\showISBNx{9798400701924}
\urldef\tempurl%
\url{https://doi.org/10.1145/3593013.3593988}
\showDOI{\tempurl}


\bibitem[Mukerjee et~al\mbox{.}(2002)]%
        {mukerjee2002}
\bibfield{author}{\bibinfo{person}{Amitabha Mukerjee}, \bibinfo{person}{Rita Biswas}, \bibinfo{person}{Kalyanmoy Deb}, {and} \bibinfo{person}{Amrit~P. Mathur}.} \bibinfo{year}{2002}\natexlab{}.
\newblock \showarticletitle{Multi{\textendash}objective Evolutionary Algorithms for the Risk{\textendash}return Trade{\textendash}off in Bank Loan Management}.
\newblock \bibinfo{journal}{\emph{International Transactions in Operational Research}} \bibinfo{volume}{9}, \bibinfo{number}{5} (\bibinfo{year}{2002}), \bibinfo{pages}{583--597}.
\newblock
\urldef\tempurl%
\url{https://doi.org/10.1111/1475-3995.00375}
\showDOI{\tempurl}
\newblock
\shownote{{\_}eprint: https://onlinelibrary.wiley.com/doi/pdf/10.1111/1475-3995.00375}.


\bibitem[Obermeyer et~al\mbox{.}(2019)]%
        {obermeyer2019}
\bibfield{author}{\bibinfo{person}{Ziad Obermeyer}, \bibinfo{person}{Brian Powers}, \bibinfo{person}{Christine Vogeli}, {and} \bibinfo{person}{Sendhil Mullainathan}.} \bibinfo{year}{2019}\natexlab{}.
\newblock \showarticletitle{Dissecting racial bias in an algorithm used to manage the health of populations}.
\newblock \bibinfo{journal}{\emph{Science}} \bibinfo{volume}{366}, \bibinfo{number}{6464} (\bibinfo{date}{10} \bibinfo{year}{2019}), \bibinfo{pages}{447--453}.
\newblock
\urldef\tempurl%
\url{https://doi.org/10.1126/science.aax2342}
\showDOI{\tempurl}
\newblock
\shownote{Publisher: American Association for the Advancement of Science}.


\bibitem[{OPEN SCIENCE COLLABORATION }(2015)]%
        {opensciencecollaboration2015}
\bibfield{author}{\bibinfo{person}{{OPEN SCIENCE COLLABORATION }}.} \bibinfo{year}{2015}\natexlab{}.
\newblock \showarticletitle{Estimating the reproducibility of psychological science}.
\newblock \bibinfo{journal}{\emph{Science}} \bibinfo{volume}{349}, \bibinfo{number}{6251} (\bibinfo{date}{08} \bibinfo{year}{2015}), \bibinfo{pages}{aac4716}.
\newblock
\urldef\tempurl%
\url{https://doi.org/10.1126/science.aac4716}
\showDOI{\tempurl}
\newblock
\shownote{Publisher: American Association for the Advancement of Science}.


\bibitem[Ortiz-Ospina and Roser(2017)]%
        {ortiz-ospina2017}
\bibfield{author}{\bibinfo{person}{Esteban Ortiz-Ospina} {and} \bibinfo{person}{Max Roser}.} \bibinfo{year}{2017}\natexlab{}.
\newblock \showarticletitle{Healthcare Spending}.
\newblock \bibinfo{journal}{\emph{Our World in Data}} (\bibinfo{date}{06} \bibinfo{year}{2017}).
\newblock
\urldef\tempurl%
\url{https://ourworldindata.org/financing-healthcare}
\showURL{%
\tempurl}


\bibitem[Pedregosa et~al\mbox{.}(2011)]%
        {pedregosa2011}
\bibfield{author}{\bibinfo{person}{F. Pedregosa}, \bibinfo{person}{G. Varoquaux}, \bibinfo{person}{A. Gramfort}, \bibinfo{person}{V. Michel}, \bibinfo{person}{B. Thirion}, \bibinfo{person}{O. Grisel}, \bibinfo{person}{M. Blondel}, \bibinfo{person}{P. Prettenhofer}, \bibinfo{person}{R. Weiss}, \bibinfo{person}{V. Dubourg}, \bibinfo{person}{J. Vanderplas}, \bibinfo{person}{A. Passos}, \bibinfo{person}{D. Cournapeau}, \bibinfo{person}{M. Brucher}, \bibinfo{person}{M. Perrot}, {and} \bibinfo{person}{E. Duchesnay}.} \bibinfo{year}{2011}\natexlab{}.
\newblock \showarticletitle{Scikit-learn: Machine Learning in Python}.
\newblock \bibinfo{journal}{\emph{Journal of Machine Learning Research}}  \bibinfo{volume}{12} (\bibinfo{year}{2011}), \bibinfo{pages}{2825{\textendash}2830}.
\newblock


\bibitem[Perrone et~al\mbox{.}(2021)]%
        {perrone2021}
\bibfield{author}{\bibinfo{person}{Valerio Perrone}, \bibinfo{person}{Michele Donini}, \bibinfo{person}{Muhammad~Bilal Zafar}, \bibinfo{person}{Robin Schmucker}, \bibinfo{person}{Krishnaram Kenthapadi}, {and} \bibinfo{person}{{Cédric} Archambeau}.} \bibinfo{year}{2021}\natexlab{}.
\newblock \showarticletitle{AIES '21: AAAI/ACM Conference on AI, Ethics, and Society}. \bibinfo{publisher}{ACM}, \bibinfo{address}{Virtual Event USA}, \bibinfo{pages}{854--863}.
\newblock
\urldef\tempurl%
\url{https://doi.org/10.1145/3461702.3462629}
\showDOI{\tempurl}


\bibitem[Pfisterer et~al\mbox{.}(2019)]%
        {pfisterer-arxiv19a}
\bibfield{author}{\bibinfo{person}{F. Pfisterer}, \bibinfo{person}{S. Coors}, \bibinfo{person}{J. Thomas}, {and} \bibinfo{person}{B. Bischl}.} \bibinfo{year}{2019}\natexlab{}.
\newblock \showarticletitle{Multi-Objective Automatic Machine Learning with AutoxgboostMC}.
\newblock \bibinfo{journal}{\emph{arXiv}}  \bibinfo{volume}{1908.10796 {[stat.ML]}} (\bibinfo{year}{2019}).
\newblock


\bibitem[Rodolfa et~al\mbox{.}(2020)]%
        {rodolfa2020}
\bibfield{author}{\bibinfo{person}{Kit~T. Rodolfa}, \bibinfo{person}{Pedro Saleiro}, {and} \bibinfo{person}{Rayid Ghani}.} \bibinfo{year}{2020}\natexlab{}.
\newblock \bibinfo{booktitle}{\emph{Bias and Fairness} (\bibinfo{edition}{2} ed.)}.
\newblock \bibinfo{publisher}{Chapman and Hall/CRC}.
\newblock
\newblock
\shownote{Num Pages: 32}.


\bibitem[Septiandri et~al\mbox{.}(2023)]%
        {septiandri2023weird}
\bibfield{author}{\bibinfo{person}{Ali~Akbar Septiandri}, \bibinfo{person}{Marios Constantinides}, \bibinfo{person}{Mohammad Tahaei}, {and} \bibinfo{person}{Daniele Quercia}.} \bibinfo{year}{2023}\natexlab{}.
\newblock \showarticletitle{{WEIRD} FAccTs: How Western, Educated, Industrialized, Rich, and Democratic is FAccT?}. In \bibinfo{booktitle}{\emph{Proceedings of the 2023 {ACM} Conference on Fairness, Accountability, and Transparency, FAccT 2023, Chicago, IL, USA, June 12-15, 2023}}. \bibinfo{publisher}{{ACM}}, \bibinfo{pages}{160--171}.
\newblock
\urldef\tempurl%
\url{https://doi.org/10.1145/3593013.3593985}
\showDOI{\tempurl}


\bibitem[Simmons et~al\mbox{.}(2011)]%
        {simmons2011}
\bibfield{author}{\bibinfo{person}{Joseph~P. Simmons}, \bibinfo{person}{Leif~D. Nelson}, {and} \bibinfo{person}{Uri Simonsohn}.} \bibinfo{year}{2011}\natexlab{}.
\newblock \showarticletitle{False-Positive Psychology: Undisclosed Flexibility in Data Collection and Analysis Allows Presenting Anything as Significant}.
\newblock \bibinfo{journal}{\emph{Psychological Science}} \bibinfo{volume}{22}, \bibinfo{number}{11} (\bibinfo{date}{11} \bibinfo{year}{2011}), \bibinfo{pages}{1359--1366}.
\newblock
\urldef\tempurl%
\url{https://doi.org/10.1177/0956797611417632}
\showDOI{\tempurl}


\bibitem[Simonsohn et~al\mbox{.}(2014)]%
        {simonsohn2014pcurve}
\bibfield{author}{\bibinfo{person}{Uri Simonsohn}, \bibinfo{person}{Leif~D. Nelson}, {and} \bibinfo{person}{Joseph~P. Simmons}.} \bibinfo{year}{2014}\natexlab{}.
\newblock \showarticletitle{P-Curve: {{A}} Key to the File-Drawer}.
\newblock \bibinfo{journal}{\emph{Journal of Experimental Psychology: General}} \bibinfo{volume}{143}, \bibinfo{number}{2} (\bibinfo{year}{2014}), \bibinfo{pages}{534--547}.
\newblock
\showISSN{1939-2222}
\urldef\tempurl%
\url{https://doi.org/10.1037/a0033242}
\showDOI{\tempurl}


\bibitem[Simonsohn et~al\mbox{.}(2020)]%
        {simonsohn2020specification}
\bibfield{author}{\bibinfo{person}{Uri Simonsohn}, \bibinfo{person}{Joseph~P. Simmons}, {and} \bibinfo{person}{Leif~D. Nelson}.} \bibinfo{year}{2020}\natexlab{}.
\newblock \showarticletitle{Specification Curve Analysis}.
\newblock \bibinfo{journal}{\emph{Nature Human Behaviour}} \bibinfo{volume}{4}, \bibinfo{number}{11} (\bibinfo{date}{Nov.} \bibinfo{year}{2020}), \bibinfo{pages}{1208--1214}.
\newblock
\showISSN{2397-3374}
\urldef\tempurl%
\url{https://doi.org/10.1038/s41562-020-0912-z}
\showDOI{\tempurl}


\bibitem[Snoek et~al\mbox{.}(2012)]%
        {snoek2012practical}
\bibfield{author}{\bibinfo{person}{Jasper Snoek}, \bibinfo{person}{Hugo Larochelle}, {and} \bibinfo{person}{Ryan~P Adams}.} \bibinfo{year}{2012}\natexlab{}.
\newblock \showarticletitle{Practical bayesian optimization of machine learning algorithms}.
\newblock \bibinfo{journal}{\emph{Advances in neural information processing systems}}  \bibinfo{volume}{25} (\bibinfo{year}{2012}).
\newblock


\bibitem[Sommers et~al\mbox{.}(2017)]%
        {sommers2017}
\bibfield{author}{\bibinfo{person}{Benjamin~D. Sommers}, \bibinfo{person}{Atul~A. Gawande}, {and} \bibinfo{person}{Katherine Baicker}.} \bibinfo{year}{2017}\natexlab{}.
\newblock \showarticletitle{Health Insurance Coverage and Health {\textemdash} What the Recent Evidence Tells Us}.
\newblock \bibinfo{journal}{\emph{New England Journal of Medicine}} \bibinfo{volume}{377}, \bibinfo{number}{6} (\bibinfo{date}{08} \bibinfo{year}{2017}), \bibinfo{pages}{586--593}.
\newblock
\urldef\tempurl%
\url{https://doi.org/10.1056/NEJMsb1706645}
\showDOI{\tempurl}


\bibitem[Steegen et~al\mbox{.}(2016)]%
        {steegen2016}
\bibfield{author}{\bibinfo{person}{Sara Steegen}, \bibinfo{person}{Francis Tuerlinckx}, \bibinfo{person}{Andrew Gelman}, {and} \bibinfo{person}{Wolf Vanpaemel}.} \bibinfo{year}{2016}\natexlab{}.
\newblock \showarticletitle{Increasing Transparency Through a Multiverse Analysis}.
\newblock \bibinfo{journal}{\emph{Perspectives on Psychological Science}} \bibinfo{volume}{11}, \bibinfo{number}{5} (\bibinfo{date}{09} \bibinfo{year}{2016}), \bibinfo{pages}{702--712}.
\newblock
\urldef\tempurl%
\url{https://doi.org/10.1177/1745691616658637}
\showDOI{\tempurl}
\newblock
\shownote{Publisher: SAGE Publications Inc}.


\bibitem[Team(2017)]%
        {pipenvmaintainerteam2017}
\bibfield{author}{\bibinfo{person}{Pipenv~Maintainer Team}.} \bibinfo{year}{2017}\natexlab{}.
\newblock \bibinfo{booktitle}{\emph{pipenv: Python Development Workflow for Humans.}}
\newblock
\urldef\tempurl%
\url{https://github.com/pypa/pipenv}
\showURL{%
\tempurl}


\bibitem[Team(2022)]%
        {rcoreteam2022}
\bibfield{author}{\bibinfo{person}{R~Core Team}.} \bibinfo{year}{2022}\natexlab{}.
\newblock \bibinfo{booktitle}{\emph{R: A Language and Environment for Statistical Computing}}.
\newblock \bibinfo{publisher}{R Foundation for Statistical Computing}, \bibinfo{address}{Vienna, Austria}.
\newblock
\urldef\tempurl%
\url{https://www.R-project.org/}
\showURL{%
\tempurl}


\bibitem[team(2020)]%
        {team2020}
\bibfield{author}{\bibinfo{person}{The pandas~development team}.} \bibinfo{year}{2020}\natexlab{}.
\newblock \bibinfo{booktitle}{\emph{pandas-dev/pandas: Pandas}}.
\newblock \bibinfo{publisher}{Zenodo}.
\newblock
\urldef\tempurl%
\url{https://doi.org/10.5281/zenodo.3509134}
\showDOI{\tempurl}
\newblock
\shownote{DOI: 10.5281/zenodo.3509134}.


\bibitem[Van~Rossum and Drake(2009)]%
        {vanrossum2009}
\bibfield{author}{\bibinfo{person}{Guido Van~Rossum} {and} \bibinfo{person}{Fred~L. Drake}.} \bibinfo{year}{2009}\natexlab{}.
\newblock \bibinfo{booktitle}{\emph{Python 3 Reference Manual}}.
\newblock \bibinfo{publisher}{CreateSpace}, \bibinfo{address}{Scotts Valley, CA}.
\newblock


\bibitem[{Watson-Daniels} et~al\mbox{.}(2023)]%
        {watson-daniels2023multitarget}
\bibfield{author}{\bibinfo{person}{Jamelle {Watson-Daniels}}, \bibinfo{person}{Solon Barocas}, \bibinfo{person}{Jake~M. Hofman}, {and} \bibinfo{person}{Alexandra Chouldechova}.} \bibinfo{year}{2023}\natexlab{}.
\newblock \showarticletitle{Multi-{{Target Multiplicity}}: {{Flexibility}} and {{Fairness}} in {{Target Specification}} under {{Resource Constraints}}}. In \bibinfo{booktitle}{\emph{Proceedings of the 2023 {{ACM Conference}} on {{Fairness}}, {{Accountability}}, and {{Transparency}}}} \emph{(\bibinfo{series}{{{FAccT}} '23})}. \bibinfo{publisher}{Association for Computing Machinery}, \bibinfo{address}{New York, NY, USA}, \bibinfo{pages}{297--311}.
\newblock
\showISBNx{9798400701924}
\urldef\tempurl%
\url{https://doi.org/10.1145/3593013.3593998}
\showDOI{\tempurl}


\bibitem[Weerts et~al\mbox{.}(2023)]%
        {weerts2023}
\bibfield{author}{\bibinfo{person}{Hilde Weerts}, \bibinfo{person}{Miroslav Dudík}, \bibinfo{person}{Richard Edgar}, \bibinfo{person}{Adrin Jalali}, \bibinfo{person}{Roman Lutz}, {and} \bibinfo{person}{Michael Madaio}.} \bibinfo{year}{2023}\natexlab{}.
\newblock \showarticletitle{Fairlearn: Assessing and Improving Fairness of AI Systems}.
\newblock \bibinfo{journal}{\emph{Journal of Machine Learning Research}} \bibinfo{volume}{24}, \bibinfo{number}{257} (\bibinfo{year}{2023}), \bibinfo{pages}{1--8}.
\newblock
\urldef\tempurl%
\url{http://jmlr.org/papers/v24/23-0389.html}
\showURL{%
\tempurl}


\bibitem[Weerts(2021)]%
        {weerts}
\bibfield{author}{\bibinfo{person}{Hilde J.~P. Weerts}.} \bibinfo{year}{2021}\natexlab{}.
\newblock \showarticletitle{An Introduction to Algorithmic Fairness}.
\newblock  (\bibinfo{year}{2021}).
\newblock
\urldef\tempurl%
\url{https://doi.org/10.48550/arXiv.2105.05595}
\showDOI{\tempurl}


\bibitem[Wickham et~al\mbox{.}(2019)]%
        {wickham2019}
\bibfield{author}{\bibinfo{person}{Hadley Wickham}, \bibinfo{person}{Mara Averick}, \bibinfo{person}{Jennifer Bryan}, \bibinfo{person}{Winston Chang}, \bibinfo{person}{Lucy McGowan}, \bibinfo{person}{Romain {François}}, \bibinfo{person}{Garrett Grolemund}, \bibinfo{person}{Alex Hayes}, \bibinfo{person}{Lionel Henry}, \bibinfo{person}{Jim Hester}, \bibinfo{person}{Max Kuhn}, \bibinfo{person}{Thomas Pedersen}, \bibinfo{person}{Evan Miller}, \bibinfo{person}{Stephan Bache}, \bibinfo{person}{Kirill {Müller}}, \bibinfo{person}{Jeroen Ooms}, \bibinfo{person}{David Robinson}, \bibinfo{person}{Dana Seidel}, \bibinfo{person}{Vitalie Spinu}, \bibinfo{person}{Kohske Takahashi}, \bibinfo{person}{Davis Vaughan}, \bibinfo{person}{Claus Wilke}, \bibinfo{person}{Kara Woo}, {and} \bibinfo{person}{Hiroaki Yutani}.} \bibinfo{year}{2019}\natexlab{}.
\newblock \bibinfo{title}{Welcome to the Tidyverse}.
\newblock
\newblock
\urldef\tempurl%
\url{https://joss.theoj.org}
\showURL{%
\tempurl}
\newblock
\shownote{DOI: 10.21105/joss.01686}.


\bibitem[Zou and Hastie(2005)]%
        {zou2005}
\bibfield{author}{\bibinfo{person}{Hui Zou} {and} \bibinfo{person}{Trevor Hastie}.} \bibinfo{year}{2005}\natexlab{}.
\newblock \showarticletitle{Regularization and Variable Selection via the Elastic Net}.
\newblock \bibinfo{journal}{\emph{Journal of the Royal Statistical Society. Series B (Statistical Methodology)}} \bibinfo{volume}{67}, \bibinfo{number}{2} (\bibinfo{year}{2005}), \bibinfo{pages}{301--320}.
\newblock
\urldef\tempurl%
\url{https://www.jstor.org/stable/3647580}
\showURL{%
\tempurl}
\newblock
\shownote{Publisher: [Royal Statistical Society, Wiley]}.


\end{thebibliography}

\newpage
\appendix

\hypertarget{supplementary-figures}{%
\section{Supplementary Figures}\label{supplementary-figures}}

\counterwithin{figure}{section}
\counterwithin{table}{section}
\renewcommand\thefigure{\thesection\arabic{figure}}
\renewcommand\thetable{\thesection\arabic{table}}

\begin{figure}[h]

{\centering \includegraphics[width=0.8\textwidth]{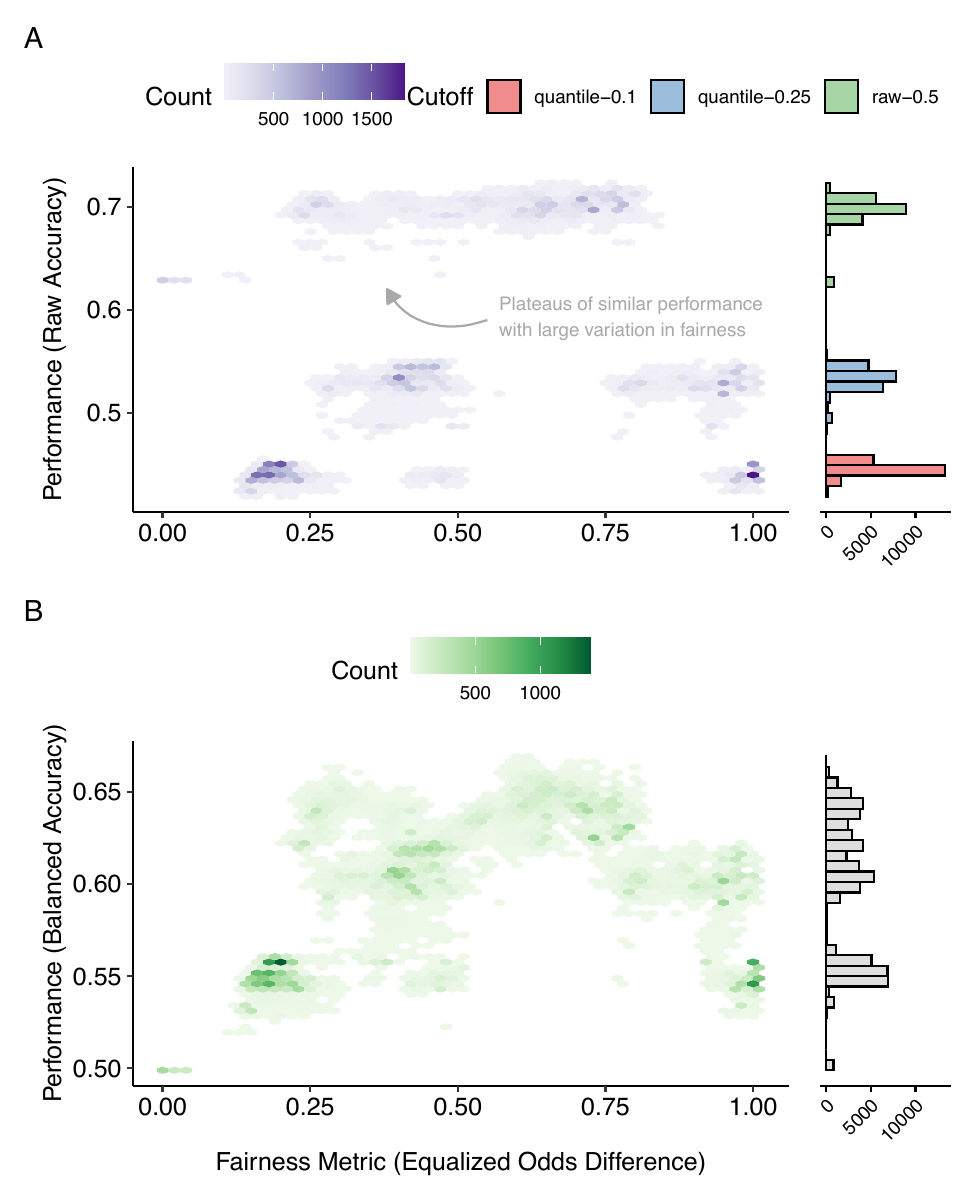}

}

\caption{\label{fig-performance-fairness-acc} \textbf{Performance and fairness are largely unrelated with clusters of low variance in performance, but high variance in fairness.} Distribution of overall performance as raw (A) or balanced (B) accuracy and fairness metric (equalized odds difference) across all multiverses. Marginal histograms show distribution of performance for different options of the \emph{Cutoff} decision in A and overall in B. A marginal histogram of the fairness metric can be seen in Figure~\ref{fig-variance}. This figure is analogous to Figure~\ref{fig-performance-fairness} in the main text.}

\end{figure}

\begin{figure}

\centering{

\includegraphics[width=0.9\textwidth]{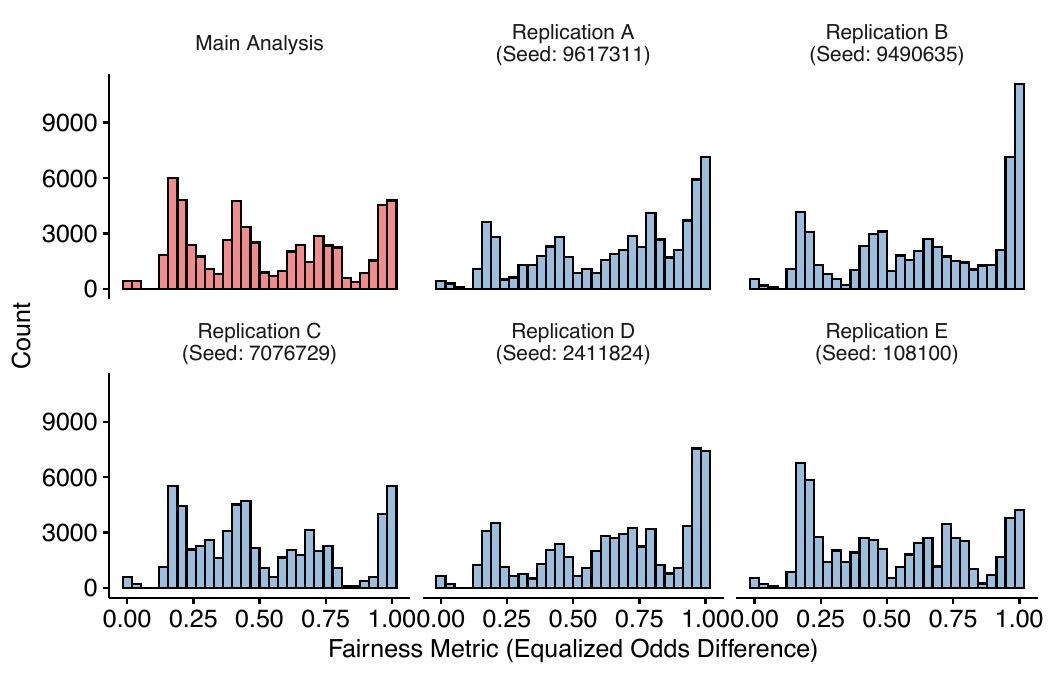}

}

\caption{\label{fig-var-replications}\textbf{Overall variation in the
multiverse is highly similar across different replications.} Distribution of
fairness metric (equalized odds difference) across universes in five
different replications alongside the results reported in the main body
of the paper. Lower values on the fairness metric indicate smaller
\emph{TPR} and \emph{FPR} differences across groups.}

\end{figure}%

\begin{figure}

\centering{

\includegraphics[width=0.9\textwidth]{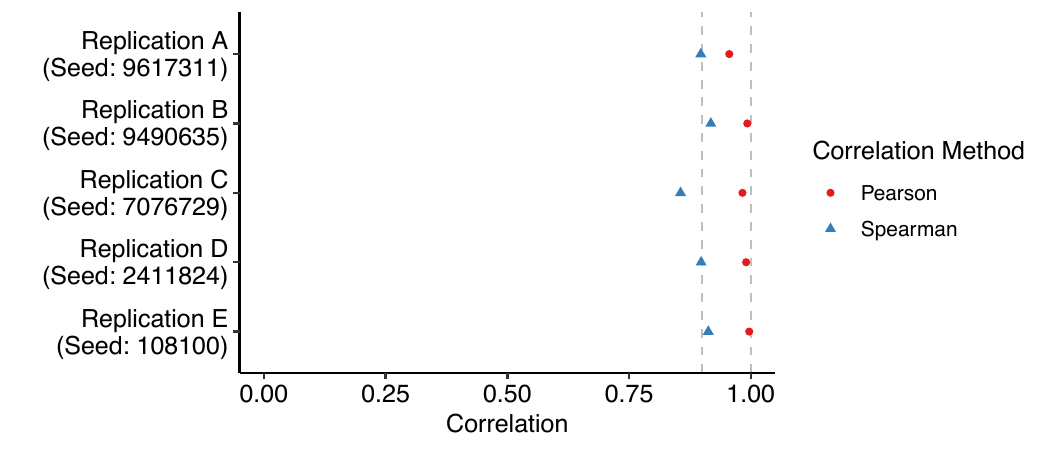}

}

\caption{\label{fig-cor-replications}\textbf{Estimates of decision
importance are similar across replications of the analysis.}
Correlations of variance decomposition / importance estimates between
the analysis reported in the main body of the paper and five
replications. Pearson correlation coefficients are consistently higher
than Spearman correlation coefficients, indicating better estimation of
high-importance decisions. Dashed lines were inserted at 0.9 and 1.0 to
indicate high correlation values.}

\end{figure}%

\begin{figure}

{\centering \includegraphics[width=0.8\textwidth]{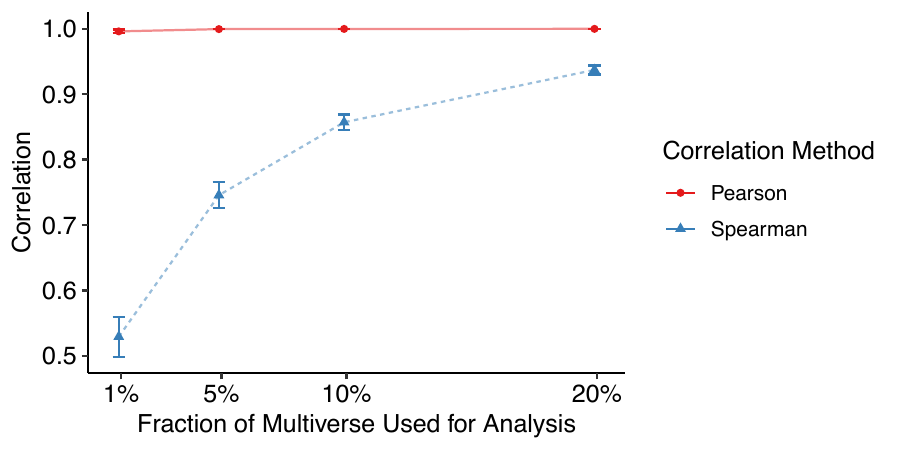}

}

\caption{\label{fig-correlations}\textbf{Conducting the analysis with
smaller subsets of the complete multiverse leads to similar results.}
Correlations of variance decomposition / importance estimates between
full dataset and random subsets of different sizes. Random subsets were
drawn 50 times with points corresponding to mean correlations and lines
to +/- 1 standard deviation. Pearson correlation coefficients are
consistently higher than Spearman correlation coefficients.}

\end{figure}

\begin{figure}

{\centering \includegraphics[width=1.0\textwidth]{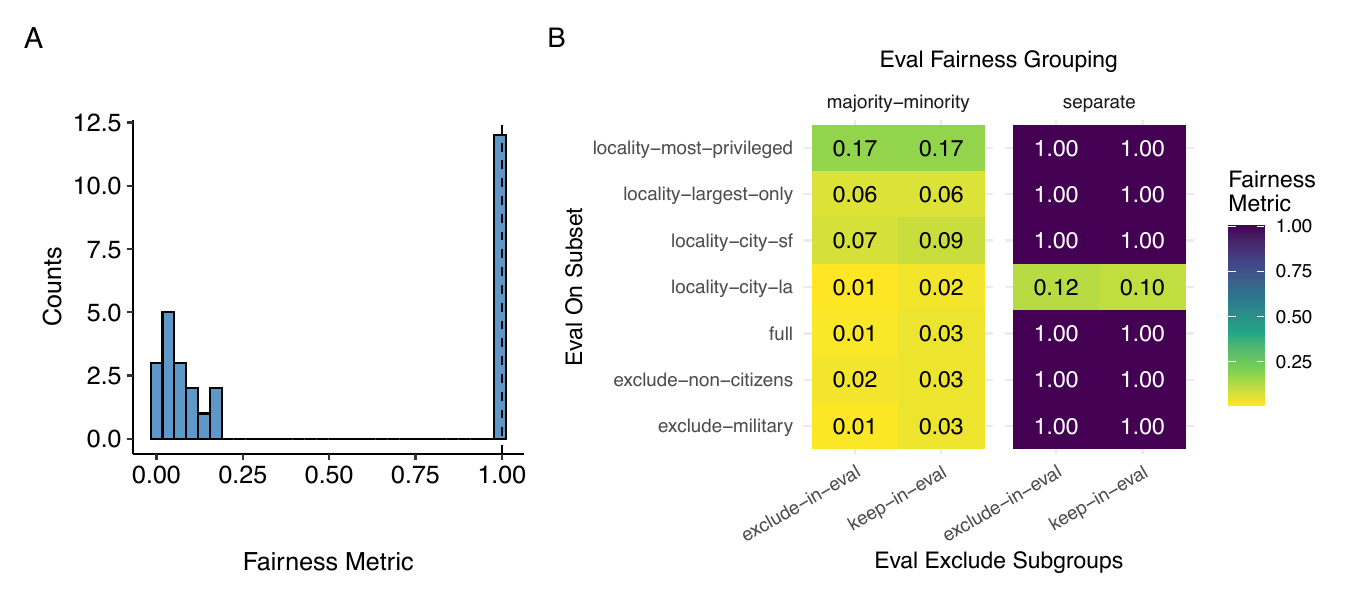}

}

\caption{\label{fig-eval-max}\textbf{Evaluation decisions can strongly interact in their effect on the fairness metric.} Overall distribution (A) and raw values (B) of the fairness metric for a single model exhibiting high variation over different decisions regarding its evaluation. This figure is analogous to Figure~\ref{fig-eval-med} in the main text.}

\end{figure}

\begin{figure}

{\centering \includegraphics[width=0.8\textwidth]{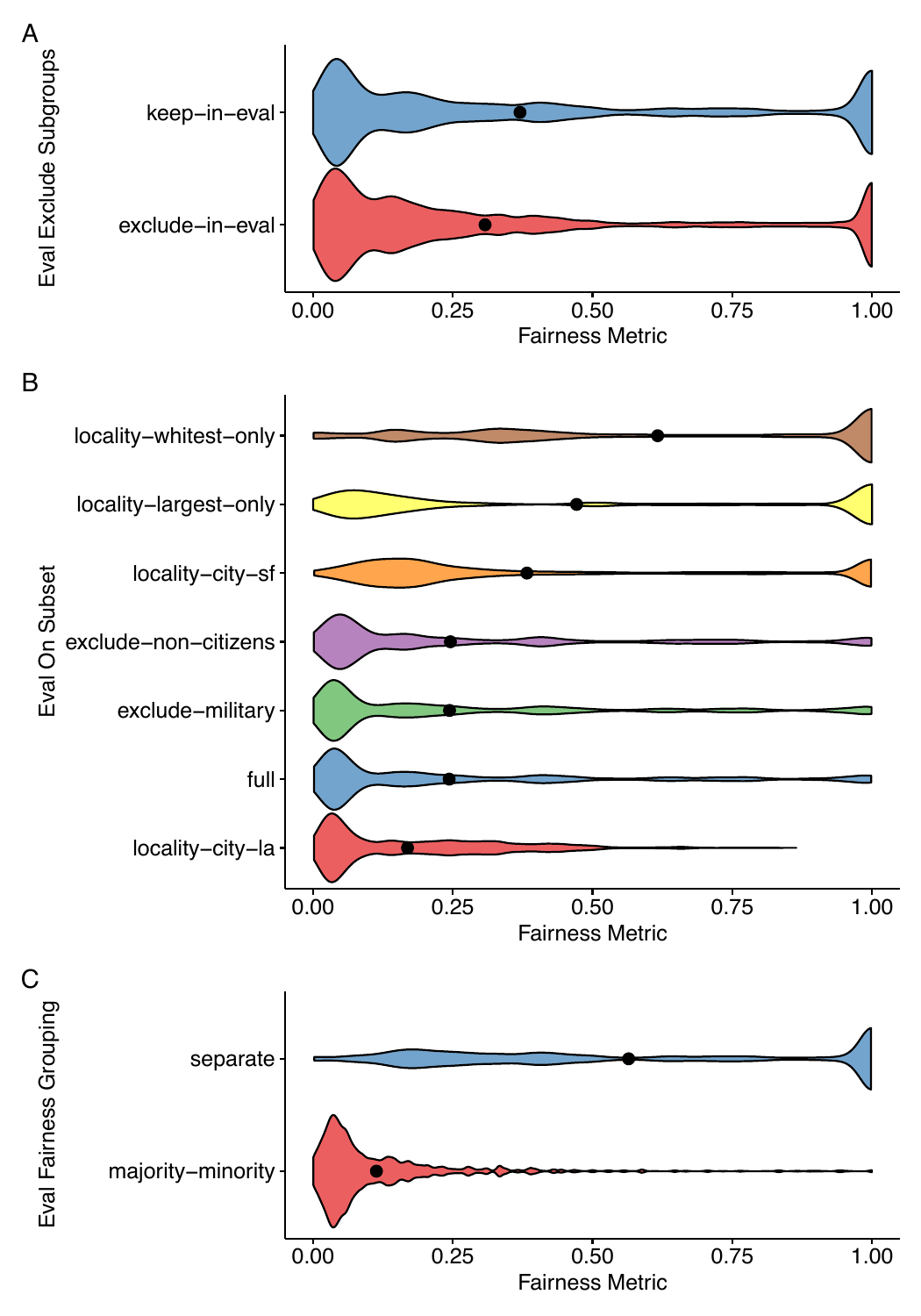}

}

\caption{\label{fig-eval-decisions}\textbf{Despite strong interactions for the same model, evaluation decisions exhibit general tendencies in how they affect algorithmic fairness.} Distribution of the fairness metric for different evaluation decisions across the complete multiverse of design decisions from studies 1 and 2.}

\end{figure}

%% begin pandoc before-bib
%% end pandoc before-bib
%% begin pandoc biblio
%% end pandoc biblio
%% begin pandoc include-after
%% end pandoc include-after
%% begin pandoc after-body
%% end pandoc after-body

\end{document}